\documentclass{article} 
\usepackage[final]{colm2025_conference}

\usepackage{microtype}
\usepackage{hyperref}
\usepackage{url}
\usepackage{booktabs}
\usepackage{subcaption}
\usepackage{lineno}
\usepackage{float}

\definecolor{darkblue}{rgb}{0, 0, 0.5}
\hypersetup{colorlinks=true, citecolor=darkblue, linkcolor=darkblue, urlcolor=darkblue}

\usepackage{pgfplots}
\usepackage{pgfplotstable}
\usetikzlibrary{pgfplots.groupplots}
\pgfplotsset{compat=1.3}
\usepackage{tikz}
\usepackage{tikz-layers}
\usepackage{longtable}
\usetikzlibrary{patterns}
\usetikzlibrary{shapes.geometric} 

\title{Zero-shot Benchmarking: A Framework for Flexible and \\ Scalable Automatic Evaluation of Language Models}

\author{
  José Pombal$^{1,2,3}$, Nuno M. Guerreiro$^{1, 2, 3, 4}$, Ricardo Rei$^{1}$ \& André F.T. Martins$^{1, 2, 3, 5}$ \\
  \ \\
  $^1$Unbabel, $^2$Instituto de Telecomunicações
  \\
  $^3$Instituto Superior Técnico, Universidade de Lisboa 
  \\
  $^4$MICS, CentraleSupélec, Université Paris-Saclay, $^5$ELLIS Unit Lisbon
  \\
    \texttt{pombal.josemaria@gmail.com}
}

\usepackage{CJKutf8}

\usepackage{hyperref}
\usepackage{url}

\usepackage{amssymb}
\usepackage{amsmath}
\usepackage{amsthm}
\usepackage{amsfonts}
\usepackage{bm}
\usepackage{nicefrac}
\usepackage{dsfont}
\usepackage{fancyvrb}
\usepackage{fvextra}
\usepackage{listings}
\usepackage{xcolor}
\usepackage{tabularx}
\usepackage[all]{nowidow}
\usepackage{xcolor}
\usepackage{colortbl}

\usepackage{booktabs}
\usepackage{multirow}
\usepackage{makecell}
\usepackage{arydshln}

\usepackage{comment}
\usepackage{xspace}
\usepackage{soul}

\usepackage{enumitem}
\usepackage[normalem]{ulem}
\setlist[itemize,enumerate]{leftmargin=*}

\makeatletter
\def\adl@drawiv#1#2#3{%
        \hskip.5\tabcolsep
        \xleaders#3{#2.5\@tempdimb #1{1}#2.5\@tempdimb}%
                #2\z@ plus1fil minus1fil\relax
        \hskip.5\tabcolsep}
\newcommand{\cdashlinelr}[1]{%
  \noalign{\vskip 2pt
           \global\let\@dashdrawstore\adl@draw
           \global\let\adl@draw\adl@drawiv}
  \cdashline{#1}[.4pt/2pt]
  \noalign{\global\let\adl@draw\@dashdrawstore
           \vskip 2pt}}
\makeatother

\definecolor{light-orange}{HTML}{fee9d4}
\definecolor{light-green}{HTML}{d8f0d3}
\definecolor{light-blue}{HTML}{dae8f5}
\definecolor{set10-red}{HTML}{e41a1c}
\definecolor{set10-blue}{HTML}{377eb8}
\definecolor{set10-green}{HTML}{4daf4a}

\definecolor{CustomBlue}{RGB}{57,83,191}
\definecolor{CustomRed}{HTML}{a75151}
\definecolor{DarkGreenOne}{RGB}{106,168,79}
\usepackage[most]{tcolorbox}

\newtcbox{\clustertab}[1]{on line, box align=base, colback={#1},colframe={#1},size=fbox,arc=2pt,top=-1.5pt, bottom=-1.5pt, left=-1.5pt, right=-1.5pt, boxrule=0pt, enlarge left by=1pt}

\newcommand{\Towervtwo}{\textsc{Tower-v2}}

\newcommand{\comet}{COMET}
\newcommand{\bleurt}{BLEURT}
\newcommand{\bleu}{\textsc{Bleu}}

\newcommand{\cometkiwi}{\textsc{CometKiwi}}
\newcommand{\xcomet}{\textsc{xComet}}
\newcommand{\gptfouro}{GPT-4o}

\newcommand{\claudethreefive}{Claude-Sonnet-3.5}

\newcommand{\metricx}{\textsc{MetricX}}

\newcommand{\autorank}{\textsc{AutoRank}}

\usepackage{pifont}
\newcommand{\cmark}{\ding{51}}%
\newcommand{\xmark}{\ding{55}}%

\begin{document}

\ifcolmsubmission
\linenumbers
\fi

\maketitle

\begin{abstract}
As language models improve and grow capable of performing more complex tasks across modalities, evaluating them automatically becomes increasingly challenging.
Developing strong and robust task-specific automatic metrics gets harder, and human-annotated test sets---which are expensive to create---saturate more quickly.
A compelling alternative is to design reliable strategies to automate the creation of test data and evaluation, but previous attempts either rely on pre-existing data, or focus solely on individual tasks.
We present Zero-shot Benchmarking (ZSB), a framework for creating high-quality benchmarks for any task by leveraging language models for both synthetic test data creation and evaluation.
ZSB is simple and flexible: it requires only the creation of a prompt for data generation and one for evaluation; it is scalable to tasks and languages where collecting real-world data is costly or impractical; it is model-agnostic, allowing the creation of increasingly challenging benchmarks as models improve.
To assess the effectiveness of our framework, we create benchmarks for five text-only tasks and a multi-modal one: general capabilities in four languages (English, Chinese, French, and Korean), translation, and general vision-language capabilities in English.
We then rank a broad range of open and closed systems on our benchmarks. 
ZSB rankings consistently correlate strongly with human rankings, outperforming widely-adopted standard benchmarks.
Through ablations, we find that strong benchmarks can be created with open models, and that judge model size and dataset variety are crucial drivers of performance.
We release all our benchmarks, and code to reproduce our experiments and to produce new benchmarks.\footnote{All data, prompts, and code we used and to create and run new benchmarks is on \href{https://github.com/deep-spin/zsb}{Github}.}
\end{abstract}

\begin{figure}
    \centering
    \includegraphics[width=\textwidth]{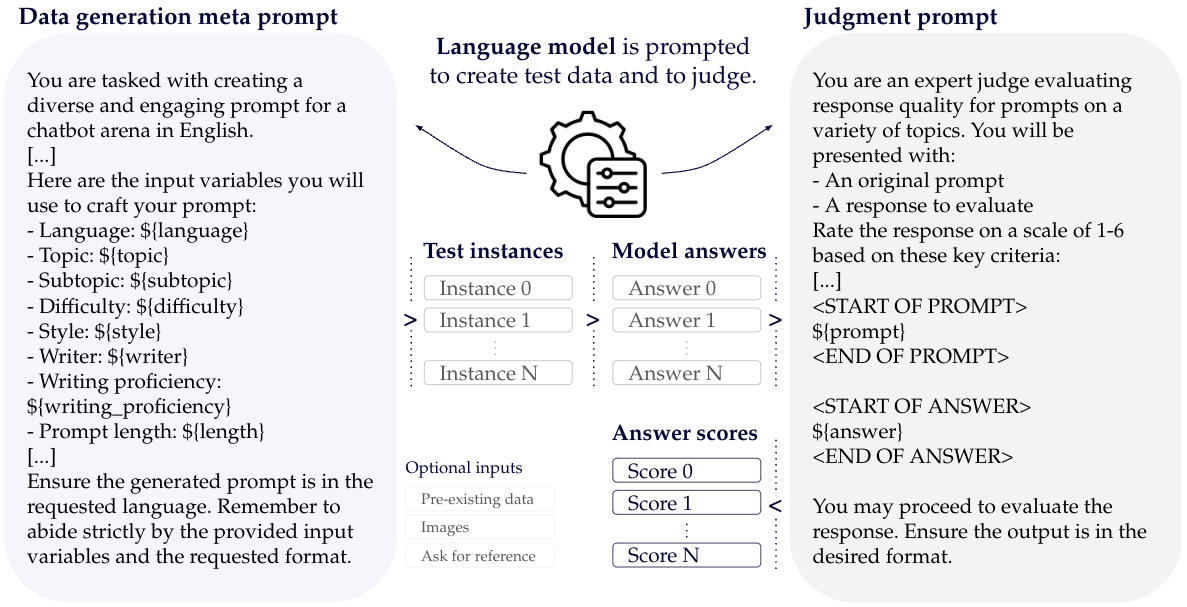}
    \caption{Zero-shot Benchmarking (ZSB) framework and task example.
    The variables in the meta prompt (inside curly brackets) allow for generating varied test instances. 
    The judgment prompt is flexible to either direct assessment or pairwise evaluation.
    The same LLM, or different ones may be used for data generation and evaluation.
    }
    \label{fig:framework}
\end{figure}

\section{Introduction}\label{sec:intro}
Automatic evaluation has played a crucial part in the rapid development of large language models (LLMs), as a proxy for more expensive and time-consuming human evaluations.
The typical approach to automatic evaluation is to create a human-annotated test set, and then evaluate a model on it using some automatic metric that may be intrinsic (e.g., accuracy on a multiple-choice test), or extrinsic (e.g., \bleurt{}~\citep{sellam2020bleurt} for text generation).
However, as the capabilities of models improve and expand to new tasks and modalities, this approach becomes challenging.
Not only because developing accurate and robust metrics for every task is hard, but also because exisiting human-annotated test sets saturate more quickly and are expensive and time-consuming to create.

To address the scarcity of automatic metrics, it has become more common to use LLMs-as-judges~\citep{zheng2023judging}, wherein an LLM is used to evaluate long-form outputs of systems~\citep{li2024leveraging,gao2024llm}.
On the other hand, to tackle the scarcity of test sets, there have been mainly two trends: 1) popular synthetically-generated multiple-choice benchmarks have emerged, such as MuSR~\citep{spraguemusr} or BoardGameQA~\citep{kazemi2024boardgameqa}; 2) efforts like the Chatbot Arena~\citep{zheng2023judging}---a widely-adopted benchmark for language model capabilities---have proposed crowdsourcing the creation of test data and evaluations.
Despite the widespread adoption of the latter approach, its limited scalability---due to the reliance on humans---constrains its usability for a wide range of tasks and for model development.
Some works propose automating the creation of evaluation data, the evaluation itself, or both.
However, they either rely on pre-existing data~\citep{li2024crowdsourced}, or focus solely on evaluating chat capabilities in English~\citep{zhao2024auto,hu2025dual}, or summarization~\citep{hu2025dual}, leaving out most real-world use-cases and tasks.

In this work, we present Zero-shot Benchmarking (ZSB), a framework for creating high-quality benchmarks for any task by leveraging language models for both synthetic test data creation and evaluation.
Our framework, showcased in Figure~\ref{fig:framework}, requires as little human effort as the creation of two prompts: \textit{i}) a meta prompt for data generation; and \textit{ii}) a judgment prompt for evaluation with a Likert scale.
ZSB unifies two highly impactful language model applications---LLM-as-a-judge and synthetic data generation---into a flexible and scalable framework for end-to-end automatic benchmarking.

ZSB represents a shift in how we approach benchmarking, particularly for specialized or emerging tasks.
Unlike traditional benchmarks that require extensive human-annotated data, ZSB can be rapidly deployed for applications where collecting real-world data would be too costly or impractical, like evaluating chat capabilities in non-English languages.
ZSB is also flexible and modular, allowing users to seamlessly create benchmarks that evaluate several skills at once, like vision-language chat capabilities, which combine textual and visual understanding.
Furthermore, contrary to static test sets, and because the framework is model-agnostic, the quality of Zero-shot benchmarks evolves alongside improvements in LLM capabilities: as more powerful models emerge, both the data generation and evaluation components can be updated simply by switching to newer models, ensuring benchmarks remain relevant and challenging.

To assess the effectiveness of ZSB, we create benchmarks for five text tasks and a multi-modal one: general capabilities\footnote{We define ``general capabilities'' in Section~\ref{sec:tasks}.} in four languages (English, Chinese, French, and Korean), translation (across eleven language pairs), and vision-language general capabilities in English.
We evaluate up to 22 LLMs and VLMs on our benchmarks, yielding rankings that correlate strongly with human rankings, often outperforming widely-adopted standard benchmarks.
We also perform a series of ablations to show that good benchmarks can be created with open models, and that judge model size and dataset variety are crucial factors for guaranteeing the best performance.
Accompanying this work, we release all our benchmarks, as well as code to reproduce our experiments and to produce new benchmarks.\footnote{Datasets and code will be released upon publication.}

\section{Zero-shot Benchmarking (ZSB)}\label{sec:framework}
ZSB (Figure~\ref{fig:framework}) is a framework for creating benchmarks fully automatically with language models, from test data generation to evaluation, for any task.
\paragraph{Data generation.} Data generation requires creating a \textbf{meta prompt}.
The meta prompt specifies the task and requests the generation of an instance that abides by a series of placeholder attributes that are relevant to the task.
For example, for chat capabilities in English, our meta prompt contains placeholders for topic, subtopic, difficulty, style, length, and writer background and proficiency, yielding a total of 9,832,320 possible combinations.
We include all attributes and their prevalence on all the released benchmarks in Appendix~\ref{apx-attrs-dists}.
We later show that variety in placeholders is important for the quality of the benchmark (\S\ref{sec:abls}).
The meta prompt may also contain a request for a reference answer, or be accompanied by pre-existing data or by images.
In order to obtain a useful data instance, it is also required to define a function for parsing the output of the meta prompt.
All meta prompts used, as well as examples of generated instances are included in Appendices~\ref{subsec:apx-meta-prompts} and~\ref{subsec:apx-examples}, respectively.

\paragraph{Evaluation.} Evaluation requires creating a \textbf{judgment prompt} that can contain a reference answer.
Contrary to previous works, we opt for a direct assessment~\citep[DA]{zheng2023judging,liu2023g,ye2023flask,kim2023prometheus,graham2018evaluation,graham2015accurate} prompt with a 6-point Likert scale, instead of a pairwise comparison prompt~\citep[PWC]{zheng2023judging,zhao2024auto,luo2024videoautoarena}.
Though DAs pose the challenge of defining an appropriate scoring scale, they are more versatile than PWC evaluation in that they do not require multiple rounds of comparisons to obtain an Elo ranking~\citep{zheng2023judging}, or the definition of a baseline answer, as done by~\citet{li2024crowdsourced}.
Furthermore, a DA prompt can leverage pre-established scoring systems, such as the MQM framework for translation~\citep{lommel2014multidimensional,burchardt-2013-multidimensional}.
We show that both DA and PWC yield system rankings of similar quality (\S\ref{sec:abls}).
All judgment prompts used are included in Appendix~\ref{subsec:apx-judge-prompts}.
In this work, we only leverage a single model for evaluation,\footnote{We test and compare several different open and closed LLMs as judges (\S\ref{sec:abls}).} but we note that ZSB can be coupled with human-in-the-loop evaluation, or approaches that attempt to mitigate the biases of a single judge, like querying multiple judges~\citep{verga2024replacing,zhao2024auto}.

\paragraph{Additional instance-level metadata.} Despite the increasing usage of synthetic benchmarks, a documentation framework is still missing, such as datasheets for datasets~\citep{gebru2021datasheets}. 
In this spirit, we provide a description and a 6-point safety score generated with \claudethreefive{} for every instance in our benchmarks.
We include the prompt we used for generating safety metadata, safety score distributions for every ZSB task, and three examples in Appendix~\ref{subsec:safety-metadata}.
The majority of instances has score greater or equal to 5 (Generally Safe), with less than 0.3\% receiving scores lower or equal to 3 (Somewhat Risky).

\section{Experimental Setup}\label{sec:exps-setup}

For each task, we create 500 test instances using \claudethreefive{}\footnote{We choose \claudethreefive{} because it is a state-of-the-art LLM, but we try other models (\S\ref{sec:abls}).} (totaling 8000 instances) and evaluate up to 22 systems.
We choose a wide variety of open and closed systems of different sizes and families; the complete list of evaluated systems per task, as well as their standings in the gold and ZSB rankings, can be found in Appendix~\ref{sec:apx-evaluated-systems}.
Evaluation is performed with \claudethreefive{} using a 6-point Likert scale in the prompt.
Systems are ranked according to their average score across all instances.

\subsection{Meta-evaluation}
Benchmarks are usually used to measure progress in the field by ranking competing systems.
As such, we evaluate ZSB by comparing its system rankings to well-established human rankings and baselines.
We use Pairwise-Accuracy~\citep[PA]{kocmi2021ship} to measure ranking correlation, which is defined as the fraction of pairs of systems for which the benchmark ranking agrees with the gold ranking.
PA is equivalent to the widely-used Kendall $\tau$ rank correlation coefficient~\citep{kendall1938new} modulo a
linear scaling and shifting~\citep{thompson2024improving}, so it will rank benchmarks in the same order as $\tau$.
We prefer to use PA for its intuitive interpretation, i.e., ``for any two systems, how likely is the benchmark to rank them in the same order as the gold ranking?''.
Following previous work~\citep{zheng2023judging,zhao2024auto}, we include Spearman's $\rho$ in Appendix~\ref{subsec:apx-spearman-correlations} (conclusions are the same as with PA).

\subsection{Tasks}\label{sec:tasks}
\paragraph{LLM General Capabilities.} The term ``general capabilities'' is often used to refer to the real-world utility of language models in addressing queries that involve core knowledge, instruction-following, and conversational capabilities~\citep{zheng2023judging}.
However, traditional test sets often rely on multiple-choice assessments, which have inherent limitations, such as lacking diversity and complexity. 
The Chatbot Arena~\citep{zheng2023judging} offers an alternative approach by crowdsourcing open-ended queries and pairwise response evaluations.
Its main advantage is that it is closer to a deployment of an LLM: humans will create diverse and complex queries as they would in a real-world setting.
Additionally, by relying on human evaluation, its rankings will reflect human preferences more than multiple-choice accuracy.
With over 1 million queries and evaluations, it has become the \textit{de facto} standard for general capabilities evaluation.
With ZSB, we want to mimic this approach, but fully automatically.
As such, we generate open-ended queries across English, Chinese, French, and Korean and evaluate up to 22 LLMs, comparing our resulting rankings with the Arena's.
We include a brief per-category analysis of the ArenaHard data and our final datasets in Appendix~\ref{sec:apx-data-analysis}.

\paragraph{Translation.} Translation is a challenging task because it is cross-lingual and open-ended , and has been at the core of language model development in recent years (e.g., it motivated the Transformer architecture~\citep{vaswani2017attention}).
Furthermore, it has a wide range of automatic evaluation metrics available that correlate well with human judgments~\citep{freitag2024llms}, which constitute strong baselines.
This provides a unique, challenging testbed for ZSB.
Thus, we create benchmarks for translation across 11 language pairs,\footnote{English-German, -Czech, -Spanish, -Russian, -Japanese, -Chinese, -Hindi, -Ukrainian, -Icelandic, Czech-Ukrainian, and Japanese-Chinese.} and evaluate up to 7 systems.
We use the WMT24 General MT Shared task~\citep{kocmi2024findings} standings as gold rankings.
The WMT shared task is a major annual competition in the field.\footnote{With over fifteen editions, WMT has become one of the biggest events at *ACL conferences.}
Each year, professional translators are hired to create high-quality test sets for a variety of language pairs.
Shared task participants then submit translations, which professional translators evaluate.
Given this reliance on professionals for test set creation and evaluation, the resulting rankings are considered the standard for comparing MT system quality.

\paragraph{VLM English General Capabilities.}
General capabilities here extend beyond text to include visual understanding, where Vision-Language Models (VLMs) have recently shown impressive performance~\citep{zhang2024vision}.
However, automatic evaluation for VLM tasks is underdeveloped compared to its text-only counterpart: high-quality benchmarks are scarce and evaluation frameworks are harder to set up and use.
Thus, it is an excellent candidate to test ZSB and its extendibility to different modalities.
We evaluate a total of 12 VLMs, and leverage gold rankings from the vision section of the Chatbot Arena.

\subsection{Baselines}\label{sec:baselines}

\paragraph{Standard Benchmarks.}
We adopt widely-used standard benchmarks as baselines~(we list all benchmarks used in Appendix~\ref{sec:apx-baselines}).
For LLM and VLM general capabilities, these benchmarks are a combination of static multiple-choice test sets.
We derive correlations with gold rankings in three different ways, which depend on how we obtain system rankings: i) by averaging the scores of each system on each test set, and then ranking systems (Average, the most common approach); ii) by taking the Borda Count~\citep{colombo2022best}, where we average the ranking of each system on each benchmark (Borda); iii) by only considering the test set in the benchmark with the highest correlations with gold rankings (Top-1).
Our goal is to establish a challenging baseline by considering the multiple ways in which practitioners might interpret benchmark results.
Note that the Borda count value for each system depends on the number of systems being considered, and that Top-1 is an oracle baseline (i.e., we know which benchmark correlates best with the gold ranking when selecting it).

The setup is slightly different for translation.
Although test sets are also static, task-specific automatic metrics are used to evaluate systems automatically.
As such, we consider the system rankings on WMT24 data yielded by the metric used by the WMT24 organizers, \autorank{}~\citep{kocmi2024findings}\footnote{To get system-level scores, \autorank{} linearly scales and averages the system-level scores (an average over the scores of all translations) of two state-of-the-art MT metrics: \metricx{}-23-XL~\citep{juraska2024metricx} (reference-based), and \cometkiwi{}-XL~\citep{rei-etal-2023-scaling} (reference-free).}, and three state-of-the-art metrics: reference-based \xcomet{}-XXL~\citep{guerreiro2023xcomet} and \metricx{}-24~\citep{juraska2024metricx}, and reference-free \cometkiwi{}-XXL~\citep{rei-etal-2023-scaling}.
These baselines are challenging, as they leverage test data created by professional translators---which also underpins the gold rankings---as well as strong, well-established MT-specific automatic metrics for evaluation.

\paragraph{M-ArenaHard Upper Bound.}
For LLM general capabilities, we consider an additional baseline: M-ArenaHard~\citep{dang2024aya}, which was obtained by translating ArenaHard~\citep{li2024crowdsourced} with Google Translate.
ArenaHard contains 500 challenging Chatbot Arena queries in English.
We use our judging framework and not the original implementation to isolate the impact of the test data's provenance on ranking performance.
We consider this to be an upper bound on the performance of our benchmarks, as the queries of ArenaHard were taken directly from the source of the gold rankings we use.

\section{Main Results}\label{sec:exps}

\begin{table}[t]
    \begin{center}
        \setlength{\tabcolsep}{3pt}
\renewcommand{\arraystretch}{1.3}
\footnotesize
\begin{tabular}{l cccc c c}
    \toprule
    & \multicolumn{4}{c}{\textbf{LLM General Capabilities}} & & \multicolumn{1}{c}{\textbf{VLM General Capabilities}} \\
    & English & Chinese & French & Korean & & English \\
    \midrule
    \multicolumn{5}{l}{\textbf{Baselines}} \\
    M-ArenaHard & \textit{0.9048} & \textit{0.9004} & 0.8333 & \textit{0.9231} & & - \\
    Std. Benchmarks (Average) & 0.8268 & 0.8269 & 0.7576 & 0.7949 & & 0.8182 \\
    Std. Benchmarks (Borda) & \textbf{0.8874} & \textbf{0.8528} & 0.7273 & 0.8205 & & 0.8182 \\
    \textcolor{gray}{Std. Benchmarks (Top-1)} & \textcolor{gray}{0.8485} & \textcolor{gray}{0.8268} & \textcolor{gray}{0.7576} & \textcolor{gray}{0.8718} & & \textcolor{gray}{0.8636} \\
    \cdashlinelr{1-7}
    \textbf{ZSB (ours)} & 0.8571 & 0.8355 & \textit{\textbf{0.8485}} & \textbf{0.8462} & & \textbf{0.8636} \\
    \bottomrule
\end{tabular}
    \end{center}
    \caption{
        Pairwise Accuracy (PA) scores for LLM and VLM general capabilities across languages.
        Bold denotes the best score apart from the upper bound (M-ArenaHard), and italics denote the best score overall. Top-1 baseline is greyed out because it is an oracle.
    }
    \label{tab:main-results-llm-vlm}
\end{table}

\subsection{LLM General Capabilities}\label{sec:exps-gen-cap}

\paragraph{ZSB outperforms standard benchmarks across all languages and is competitive with test sets obtained by curating human data.}

Table~\ref{tab:main-results-llm-vlm} shows the PA scores for each language.
ZSB outperforms averaging test sets in standard benchmarks---one of the most used method for evaluating general capabilities---across the board.
Furthermore, it outperforms Borda count---the strongest baseline---on French and Korean.
The same trend is observed when comparing against almost all individual benchmarks, instead of aggregations (see Appendix~\ref{sec:apx-baselines-llm}).
Remarkably, ZSB outperforms the upper bound in French at 0.845 PA.
These findings outline the potential of ZSB for creating high-quality multilingual benchmarks at a low cost: our English test set cost roughly \$5, while ArenaHard~\citep{li2024crowdsourced} cost \$500, and GPQA~\citep{rein2023gpqa}---which got 0.73 PA---cost upward of \$100,000~\citep{zheng2023judging}.

\begin{table}[t]
    \begin{center}
        \setlength{\tabcolsep}{3pt}
\renewcommand{\arraystretch}{1.3}
\footnotesize
\begin{tabular}{l cccc c c}
\toprule
& \multicolumn{4}{c}{\textbf{WMT24 Test Data}} & & \textbf{ZSB (ours)} \\
& \metricx{}-24-XXL & \xcomet{}-XXL & \cometkiwi{}-XXL & \autorank{} & & LLM Judge \\
\midrule
\textbf{Translation} & 0.8322 & 0.8322 & 0.8112 & 0.8182 & & 0.7902 \\
\bottomrule
\end{tabular}
    \end{center}
    \caption{
        Pairwise Accuracy (PA) scores for translation. We consider up to 7 systems across 11 language pairs, totalling 143 system comparisons.
    }
    \label{tab:main-results-mt}
\end{table}

\subsection{Translation}\label{sec:exps-mt}

\paragraph{ZSB is competitive with combining human test data and MT-specific automatic metrics.}
Table~\ref{tab:main-results-mt} shows PA scores for MT, aggregated across language pairs (we show by-LP results in Appendix~\ref{apx-per-lp-results}).
At 0.79 PA, ZSB is only two points below \autorank{} and \cometkiwi{}-XXL, the state-of-the-art reference-free metric for MT.
ZSB is also close to reference-based \xcomet{} and \metricx{}. 
This is a remarkable result given the complexity of the task and the quality of the baseline: not only are the metrics we report tailored for MT, but the data underlying their system rankings is the same as the gold rankings.
Because of the high cost of hiring professionals to create data and evaluate systems, WMT lacks language coverage and many flavours of translation (e.g., transcreation, which mixes translation with rewriting for cultural adaptation).
ZSB could potentially be used for creating benchmarks for any language and task (or mixture thereof) with minimal human effort, which may alleviate this issue.

\subsection{VLM English General Capabilities}\label{sec:exps-vlm-gen-cap}
\paragraph{Data creation.}
For each instance of this task, we randomly sample an image---and only the image---from the validation set of \texttt{textvqa}~\citep{singh2019towards}, a question-answering dataset with a wide variety of images.
The meta prompt then queries the model to generate a prompt related to the image in the same style of the LLM general capabilities task, but without attribute placeholders.
An interesting direction for future work would be creating a test set from synthetically-generated images.

\paragraph{ZSB outperforms multimodal standard benchmarks.} Table~\ref{tab:main-results-llm-vlm} shows that ZSB outperforms standard benchmarks by a considerable margin---4.5 PA points---regardless of the aggregation method.
This is a strong result, considering that the leaderboard leverages 8 datasets, while our dataset contains only 500 instances.
Notably, all but one test set, MMMU, are outperformed by ZSB (see Table~\ref{tab:llm-vlm-baselines} in the Appendix).
This highlights the flexibility of Zero-shot Benchmarking in seamlessly extending to multimodal settings.

\section{Ablations}\label{sec:abls}

\begin{table}[t]
    \begin{center}
        \setlength{\tabcolsep}{3pt}
\renewcommand{\arraystretch}{1.3}
\footnotesize
\begin{tabular}{l ccccccc}
    \toprule
    & \multicolumn{7}{c}{\textbf{Judge}} \\
    \textbf{Data Generator} & Claude & Llama-70B & Qwen-3B & Qwen-7B & Qwen-14B & Qwen-32B & Qwen-72B \\
    \midrule	
    Claude & \textbf{\textit{0.8528}}* & 0.8182 & \textbf{0.7446} & 0.7013 & 0.7100 & 0.8139 & 0.7792 \\
    Llama-70B & 0.7532 & \textit{0.7792} & 0.6277 & 0.6797 & 0.7273 & 0.7706 & 0.7619 \\
    Qwen-3B & 0.7706 & \textit{0.7792}& 0.6840 & 0.6623 & 0.7403 & 0.7749 & 0.7662 \\
    Qwen-7B & \textit{0.7835} & 0.7749 & 0.6753 & 0.6667 & 0.6970 & 0.7576 & 0.7489 \\
    Qwen-14B & \textit{0.8052} & 0.7922 & 0.7143 & \textbf{0.7273} & 0.7446 & 0.7749 & 0.7965 \\
    Qwen-32B & \textit{0.7922} & \textit{0.7922} & 0.6537 & 0.7100 & \textbf{0.7922} & 0.7619 & 0.8139 \\
    Qwen-72B & 0.7922 & \textbf{\textit{0.8268}} & 0.6797 & 0.7100 & 0.7446 & \textbf{0.8312} & \textbf{0.8182} \\
    \bottomrule
\end{tabular}

    \end{center}
    \caption{
        Pairwise accuracy of Zero-shot Benchmark for general capabilities in English with various data generation and judge models. 
        Italic denotes the best judge for a given generator; bold the best data generator for a given judge. 
        The * denotes our original configuration.
    }
    \label{tab:gen-judge-ablation}
\end{table}

We ablate a series of components in our framework and experimental setup to understand their impact on the quality of the ZSB text-only general capabilities benchmarks.

\paragraph{Open models can be used to generate high-quality benchmark data.} The vast majority benchmarks comprised of synthetic data were generated with proprietary models~\citep{Fan2023LargeLM,spraguemusr,wu2024unigen}.
Using open models for data generation is an attractive alternative for three reasons: \textit{i)} data may be cheaper to generate; \textit{ii)} the resulting data can be free of usage restrictions; and, perhaps most importantly, \textit{iii)} it is easier to improve benchmarks as ZSB improves in tandem with open model advancements.
Thus, we evaluate the performance of open models Llama-3.3-70B-Instruct~\citep{grattafiori2024llama} and Qwen2.5-\{3, 7, 14, 32, 72\}B-Instruct~\citep{yang2024qwen2} on both the data generation and judgment components of our framework.
Table~\ref{tab:gen-judge-ablation} shows that benchmarks created with Llama-3.3-70B-Instruct or Qwen2.5-72B-Instruct can achieve strong correlations with human rankings in English, and, remarkably, outside of English (see Appendix~\ref{subsec:apx-data-judge-ablations}).
Importantly, Qwen outperforms averaging standard benchmarks across the board.
Choosing the right model is crucial for the quality of the benchmark---for example, Qwen is especially strong in Korean, while Llama lags behind.
Finally, to further illustrate that the quality of ZSB can improve alongside that of open models, we present results for general capabilities using Llama 4 Maverick~\citep{meta2025llama} as a data generator and judge in Appendix~\ref{subsec:apx-maverick}, showing it outperforms Llama 3.3 on 3 out of 4 languages.

\paragraph{Open models are strong judges but they must be somewhat large.} Table~\ref{tab:gen-judge-ablation} also shows the performance of different judge models, given the same test data.
Though smaller Qwen models perform worse in general, high performance can be achieved from 14B parameters onwards, especially for Chinese and Korean (see Appendix~\ref{subsec:apx-data-judge-ablations}).
Notably, using Llama-3.3-70B-Instruct as a judge with data generated by \claudethreefive{} leads to better ranking correlations on Chinese and French, even outperforming M-ArenaHard on the latter (see Appendix~\ref{subsec:apx-data-judge-ablations}).
It is also viable to use open models on the full ZSB stack, i.e., both data generation and evaluation: pairing Qwen and Llama, or Qwen models models can lead to higher PA outside of English than exclusively using Claude.

\paragraph{Judges perform better on data created by models of the same family.} 
An interesting trend from Table~\ref{tab:gen-judge-ablation} is that Qwen judges tend to perform better on data generated by other Qwen models.
For example, from sizes 7B to 72B, Qwen judges often perform better on Qwen-generated data than on data generated by Claude, otherwise the best data generator.
A possible explanation is that models are more likely to generate judgments in the desired format when they are used on data generated by their family (see Table~\ref{tab:judge-pct-failures} in the Appendix).
While in this work we focused on Claude, this shows that there are performance gains in further tailouring the data generation and judging processes to open models.

\begin{table}[t]
    \begin{center}
        \setlength{\tabcolsep}{3pt}
\renewcommand{\arraystretch}{1.3}
\footnotesize
\begin{tabular}{l ccc c cc}
\toprule
& \multicolumn{3}{c}{\textbf{Claude}} & & \multicolumn{2}{c}{\textbf{Llama}} \\
& DA & PWC w/ Baseline & ArenaHard Elo & & DA & PWC All \\
\midrule
\textbf{Pairwise Accuracy} & \textbf{0.8528}* & 0.8268 & 0.8442 & & 0.8355 & 0.8139 \\
\bottomrule
\end{tabular}
    \end{center}
    \caption{
        Pairwise-accuracy of ZSB for general capabilities in English with various judgment approaches.
        The * denotes our original configuration.
        In ``PWC w/ Baseline'', the final score of each system is the win rate against \gptfouro{}.
        In ``ArenaHard Elo'', we employ the approach of~\citet{li2024crowdsourced} to compute an Elo ranking, using \claudethreefive{} generations as the baseline answers with PWC evaluation.
        In ``PWC All'', we compare all systems against each other, and obtain an Elo ranking from those battles.
        In the latter, we employ Llama-3.3-70B-Instruct as a judge instead of \claudethreefive{} for cost reasons.
    }
    \label{tab:judge-ablation}
\end{table}

\paragraph{Direct assessment is a viable alternative to pairwise judging.} In this work, we opted for a direct assessment judgment approach instead of a pairwise one.
DA makes it easier to add new models to a leaderboard, and can be integrated with existing task-specific scoring frameworks.
However, to the best of our knowledge, this type of judging had not been employed before, as past works use forms of PWC to obtain Elo scores for ranking systems~\citep{zheng2023judging,zhao2024auto}.
In Table~\ref{tab:judge-ablation}, we show that the performance of DA and PWC is comparable in English, with DA slightly outperforming PWC among all systems, or PWC against a baseline answer generated by \claudethreefive{} (the method for computing Elo rankings used in ArenaHardAuto~\citep{zheng2023judging}).

\paragraph{Data instances adhere to meta-prompt specifications, and benchmark performance is robust to meta-prompt variations, but reference answer quality is contentious.} Throughout this work, we only used a single meta-prompt for creating general capabilities data, and we mostly performed reference-less evaluation.
However, one step to ensure the practical utility and versatility of our framework is to verify its reliability with respect to the reference answer it provides and to the meta-prompt---both in terms of whether the generated instances respect it, and whether useful instances are still generated after modifying the prompt lexically.
Thus, in Table~\ref{tab:data-verifications} we present five statistics for all the languages we consider: (i) the percentage of data instances that abide by the meta-prompt's attributes; (ii) the percentage of perfect references (scored 1 to 6); (iii) the percentage of score 5 references; (iv) and (v) the PA average and standard deviation of our benchmarks generated with three meta-prompts (the default prompt plus two rewrites).\footnote{We prompt \claudethreefive{} to obtain all statistics; the scripts and prompts used are in the Github repository linked at the beginning of this work.}
The statistics show similar patterns across languages. 
While the generated data is consistent with meta prompt specifications, reference quality could be improved: the vast majority of references are quite good (score 5), but only a small fraction are perfect.

\begin{table}[t]
    \begin{center}
    \setlength{\tabcolsep}{3pt}
\renewcommand{\arraystretch}{1.3}
\footnotesize
\begin{tabular}{l cccc}
    \toprule
    & \multicolumn{4}{c}{\textbf{LLM General Capabilities}} \\
    & English & Chinese & French & Korean \\
    \midrule
    \multicolumn{5}{l}{\textbf{Baselines}} \\
    \% consistency w/ meta prompt attributes & 95.8 & 97.4 & 99.6 & 99.0  \\ 
    \% perfect references & 7.4 & 26.8 & 5.2  & 2.0 \\ 
    \% score 5 references & 89.2 & 68.6 & 86.6 & 88.2 \\ 
    \% Avg. PA for 3 meta prompts & 0.8470 & 0.8341 & 0.8283 & 0.8419 \\ 
    \% Std. dev. of PA for 3 meta prompts & 0.0090 & 0.0025 & 0.0175 & 0.0074 \\ 
    \bottomrule
\end{tabular}
    \end{center}
    \caption{
        Data statistics towards ensuring the practical robustness and versatility of our data generation framework: (i) the percentage of data instances that abide by the meta-prompt's attributes; (ii) the percentage of perfect references (scored 1 to 6); (iii) the percentage of score 5 references; (iv) and (v) the PA average and standard deviation of our benchmarks generated with three meta-prompts (the default prompt plus two rewrites).
    }
    \label{tab:data-verifications}
\end{table}

\paragraph{The importance of dataset size is limited beyond 100 instances.} Figure~\ref{fig:size-ablation} shows the PA scores of our English benchmark over different dataset sizes.
While performance increases substantially in the first 100 instances, it stagnates after that, only reaching 2.5 extra PA points at 500 instances.
This finding hints at the limited importance of dataset size, but it is encouraging in that small datasets are more accessible and easier to create and curate.

\definecolor{CustomBlue}{RGB}{0, 114, 189}
\definecolor{battleshipgrey}{rgb}{0.3, 0.3, 0.3}
\definecolor{brilliantrose}{rgb}{1.0, 0.33, 0.64}
\definecolor{americanrose}{rgb}{1.0, 0.01, 0.24}
\definecolor{jweigreen}{rgb}{0,0.45,0.24}
\definecolor{bluegray}{rgb}{0.1, 0.1, 0.4}
\definecolor{ao(english)}{rgb}{0.0, 0.5, 0.0}
\definecolor{blanchedalmond}{rgb}{1.0, 0.92, 0.8}
\definecolor{atomictangerine}{rgb}{1.0, 0.6, 0.4}
\definecolor{chocolate(web)}{rgb}{0.82, 0.41, 0.12}
\definecolor{bananayellow}{rgb}{1.0, 0.88, 0.21}
\definecolor{goldenbrown}{rgb}{0.6, 0.4, 0.08}
\definecolor{aliceblue}{rgb}{0.94, 0.97, 1.0}
\definecolor{beige}{rgb}{0.96, 0.96, 0.86}
\definecolor{babyblue}{rgb}{0.54, 0.81, 0.94}
\definecolor{camel}{rgb}{0.76, 0.6, 0.42}
\definecolor{cinnamon}{rgb}{0.82, 0.41, 0.12}
\definecolor{deepskyblue}{rgb}{0.0, 0.75, 1.0}
\definecolor{frenchblue}{rgb}{0.0, 0.45, 0.73}
\definecolor{classicrose}{rgb}{0.98, 0.8, 0.91}
\definecolor{frenchrose}{rgb}{0.96, 0.29, 0.54}
\definecolor{frenchlilac}{rgb}{0.53, 0.38, 0.56}
\definecolor{frenchbeige}{rgb}{0.65, 0.48, 0.36}

\begin{figure}[t]
    \begin{subfigure}[b]{0.48\textwidth}
        \centering
        \pgfplotsset{
            footnotesize,
            samples=10,
            xmin=-1, xmax=501,
            ymin=0.6, ymax=0.87,
            xtick={0,100,200,300,400,500},
            xticklabels={0,100,200,300,400,500},
            ytick={0.5, 0.55, 0.60, 0.65, 0.70, 0.75, 0.80, 0.85},
            yticklabels={0.5, 0.55, 0.60, 0.65, 0.70, 0.75, 0.80, 0.85},
            grid style={dashed},
            grid=both,
            xlabel=Number of instances in test set,
            x label style={at={(axis description cs:0.5,-0.15)},anchor=north},
        }
        
        \begin{tikzpicture}
        \begin{groupplot}[
            group style = {group size = 1 by 1, horizontal sep = 24pt},
            width = \textwidth, 
            height = 4cm
        ]
            \nextgroupplot[
                align = center,
                legend style={at={(1.,-0.6)},anchor=south
                ,/tikz/column 2/.style={column sep=8pt,}
                ,/tikz/column 4/.style={column sep=8pt,}
                ,/tikz/column 6/.style={column sep=8pt,}
                ,/tikz/column 8/.style={column sep=8pt,}
                ,/tikz/column 10/.style={column sep=8pt,}
                ,},
                legend cell align={left},
                legend style={draw=none},
                legend columns=3,
                axis x line*=bottom,
                axis y line*=left,
                ylabel=Pairwise Accuracy,
                y label style={at={(axis description cs:-0.15,0.485)},anchor=south},
                xtick pos=bottom,
                ytick pos=left,
            ]
            \addplot[ %
            color=CustomBlue, line width=1.1pt,
        ]
        coordinates {(1, 0.6926) (2, 0.684) (3, 0.6926) (4, 0.7186) (5, 0.7273) (6, 0.7316) (7, 0.7446) (8, 0.7489) (9, 0.7576) (10, 0.7835) (11, 0.7749) (12, 0.7835) (13, 0.7792) (14, 0.7835) (15, 0.7792) (16, 0.7749) (17, 0.7922) (18, 0.7965) (19, 0.7965) (20, 0.7965) (21, 0.8182) (22, 0.8052) (23, 0.8139) (24, 0.8139) (25, 0.8095) (26, 0.8268) (27, 0.8312) (28, 0.8312) (29, 0.8268) (30, 0.8225) (31, 0.8268) (32, 0.8225) (33, 0.8225) (34, 0.8268) (35, 0.8268) (36, 0.8312) (37, 0.8355) (38, 0.8355) (39, 0.8355) (40, 0.8182) (41, 0.8225) (42, 0.8225) (43, 0.8182) (44, 0.8052) (45, 0.7965) (46, 0.7965) (47, 0.8052) (48, 0.8268) (49, 0.8225) (50, 0.8312) (51, 0.8312) (52, 0.8398) (53, 0.8485) (54, 0.8485) (55, 0.8485) (56, 0.8485) (57, 0.8485) (58, 0.8442) (59, 0.8442) (60, 0.8442) (61, 0.8442) (62, 0.8442) (63, 0.8398) (64, 0.8398) (65, 0.8442) (66, 0.8442) (67, 0.8398) (68, 0.8442) (69, 0.8442) (70, 0.8442) (71, 0.8485) (72, 0.8442) (73, 0.8442) (74, 0.8442) (75, 0.8442) (76, 0.8442) (77, 0.8355) (78, 0.8355) (79, 0.8398) (80, 0.8355) (81, 0.8355) (82, 0.8398) (83, 0.8398) (84, 0.8442) (85, 0.8442) (86, 0.8442) (87, 0.8442) (88, 0.8398) (89, 0.8398) (90, 0.8355) (91, 0.8398) (92, 0.8312) (93, 0.8312) (94, 0.8312) (95, 0.8312) (96, 0.8312) (97, 0.8268) (98, 0.8268) (99, 0.8268) (100, 0.8268) (101, 0.8398) (102, 0.8442) (103, 0.8442) (104, 0.8442) (105, 0.8398) (106, 0.8485) (107, 0.8355) (108, 0.8355) (109, 0.8355) (110, 0.8355) (111, 0.8355) (112, 0.8355) (113, 0.8355) (114, 0.8355) (115, 0.8355) (116, 0.8355) (117, 0.8355) (118, 0.8355) (119, 0.8355) (120, 0.8355) (121, 0.8398) (122, 0.8398) (123, 0.8442) (124, 0.8442) (125, 0.8442) (126, 0.8442) (127, 0.8442) (128, 0.8442) (129, 0.8442) (130, 0.8442) (131, 0.8398) (132, 0.8398) (133, 0.8398) (134, 0.8398) (135, 0.8398) (136, 0.8398) (137, 0.8398) (138, 0.8398) (139, 0.8442) (140, 0.8442) (141, 0.8442) (142, 0.8442) (143, 0.8442) (144, 0.8442) (145, 0.8442) (146, 0.8442) (147, 0.8398) (148, 0.8398) (149, 0.8312) (150, 0.8312) (151, 0.8398) (152, 0.8398) (153, 0.8398) (154, 0.8398) (155, 0.8442) (156, 0.8442) (157, 0.8442) (158, 0.8442) (159, 0.8442) (160, 0.8398) (161, 0.8398) (162, 0.8398) (163, 0.8442) (164, 0.8442) (165, 0.8442) (166, 0.8442) (167, 0.8398) (168, 0.8398) (169, 0.8398) (170, 0.8398) (171, 0.8398) (172, 0.8398) (173, 0.8398) (174, 0.8398) (175, 0.8442) (176, 0.8442) (177, 0.8442) (178, 0.8442) (179, 0.8442) (180, 0.8442) (181, 0.8442) (182, 0.8442) (183, 0.8398) (184, 0.8398) (185, 0.8442) (186, 0.8442) (187, 0.8442) (188, 0.8442) (189, 0.8442) (190, 0.8442) (191, 0.8442) (192, 0.8442) (193, 0.8442) (194, 0.8442) (195, 0.8442) (196, 0.8442) (197, 0.8442) (198, 0.8442) (199, 0.8442) (200, 0.8442) (201, 0.8442) (202, 0.8442) (203, 0.8442) (204, 0.8442) (205, 0.8442) (206, 0.8442) (207, 0.8398) (208, 0.8398) (209, 0.8398) (210, 0.8398) (211, 0.8398) (212, 0.8398) (213, 0.8398) (214, 0.8398) (215, 0.8398) (216, 0.8398) (217, 0.8398) (218, 0.8355) (219, 0.8398) (220, 0.8398) (221, 0.8398) (222, 0.8355) (223, 0.8355) (224, 0.8355) (225, 0.8355) (226, 0.8355) (227, 0.8355) (228, 0.8355) (229, 0.8355) (230, 0.8355) (231, 0.8355) (232, 0.8355) (233, 0.8355) (234, 0.8355) (235, 0.8355) (236, 0.8355) (237, 0.8355) (238, 0.8355) (239, 0.8355) (240, 0.8355) (241, 0.8355) (242, 0.8355) (243, 0.8355) (244, 0.8355) (245, 0.8355) (246, 0.8355) (247, 0.8355) (248, 0.8355) (249, 0.8355) (250, 0.8355) (251, 0.8312) (252, 0.8355) (253, 0.8355) (254, 0.8355) (255, 0.8355) (256, 0.8355) (257, 0.8355) (258, 0.8398) (259, 0.8398) (260, 0.8355) (261, 0.8355) (262, 0.8355) (263, 0.8355) (264, 0.8355) (265, 0.8355) (266, 0.8355) (267, 0.8355) (268, 0.8355) (269, 0.8355) (270, 0.8355) (271, 0.8355) (272, 0.8355) (273, 0.8355) (274, 0.8355) (275, 0.8355) (276, 0.8355) (277, 0.8355) (278, 0.8398) (279, 0.8398) (280, 0.8398) (281, 0.8398) (282, 0.8398) (283, 0.8398) (284, 0.8398) (285, 0.8398) (286, 0.8398) (287, 0.8355) (288, 0.8355) (289, 0.8355) (290, 0.8355) (291, 0.8355) (292, 0.8355) (293, 0.8355) (294, 0.8355) (295, 0.8355) (296, 0.8355) (297, 0.8442) (298, 0.8485) (299, 0.8442) (300, 0.8398) (301, 0.8398) (302, 0.8398) (303, 0.8398) (304, 0.8398) (305, 0.8355) (306, 0.8355) (307, 0.8355) (308, 0.8355) (309, 0.8398) (310, 0.8398) (311, 0.8355) (312, 0.8355) (313, 0.8355) (314, 0.8355) (315, 0.8398) (316, 0.8398) (317, 0.8398) (318, 0.8442) (319, 0.8442) (320, 0.8442) (321, 0.8485) (322, 0.8485) (323, 0.8485) (324, 0.8442) (325, 0.8355) (326, 0.8355) (327, 0.8355) (328, 0.8355) (329, 0.8398) (330, 0.8398) (331, 0.8398) (332, 0.8398) (333, 0.8442) (334, 0.8485) (335, 0.8485) (336, 0.8442) (337, 0.8442) (338, 0.8442) (339, 0.8398) (340, 0.8442) (341, 0.8442) (342, 0.8442) (343, 0.8442) (344, 0.8442) (345, 0.8442) (346, 0.8442) (347, 0.8442) (348, 0.8442) (349, 0.8442) (350, 0.8485) (351, 0.8485) (352, 0.8485) (353, 0.8485) (354, 0.8485) (355, 0.8485) (356, 0.8485) (357, 0.8485) (358, 0.8485) (359, 0.8485) (360, 0.8485) (361, 0.8485) (362, 0.8485) (363, 0.8485) (364, 0.8485) (365, 0.8485) (366, 0.8485) (367, 0.8485) (368, 0.8485) (369, 0.8485) (370, 0.8485) (371, 0.8528) (372, 0.8528) (373, 0.8528) (374, 0.8528) (375, 0.8528) (376, 0.8528) (377, 0.8528) (378, 0.8485) (379, 0.8528) (380, 0.8528) (381, 0.8528) (382, 0.8528) (383, 0.8528) (384, 0.8528) (385, 0.8528) (386, 0.8528) (387, 0.8528) (388, 0.8528) (389, 0.8528) (390, 0.8528) (391, 0.8528) (392, 0.8528) (393, 0.8528) (394, 0.8528) (395, 0.8528) (396, 0.8528) (397, 0.8528) (398, 0.8528) (399, 0.8528) (400, 0.8528) (401, 0.8528) (402, 0.8528) (403, 0.8528) (404, 0.8528) (405, 0.8528) (406, 0.8528) (407, 0.8528) (408, 0.8528) (409, 0.8528) (410, 0.8528) (411, 0.8528) (412, 0.8528) (413, 0.8528) (414, 0.8528) (415, 0.8528) (416, 0.8528) (417, 0.8528) (418, 0.8528) (419, 0.8528) (420, 0.8528) (421, 0.8528) (422, 0.8528) (423, 0.8528) (424, 0.8528) (425, 0.8528) (426, 0.8528) (427, 0.8528) (428, 0.8528) (429, 0.8528) (430, 0.8528) (431, 0.8528) (432, 0.8528) (433, 0.8528) (434, 0.8528) (435, 0.8528) (436, 0.8528) (437, 0.8528) (438, 0.8528) (439, 0.8528) (440, 0.8528) (441, 0.8528) (442, 0.8528) (443, 0.8528) (444, 0.8528) (445, 0.8528) (446, 0.8528) (447, 0.8528) (448, 0.8528) (449, 0.8528) (450, 0.8528) (451, 0.8528) (452, 0.8528) (453, 0.8528) (454, 0.8571) (455, 0.8571) (456, 0.8571) (457, 0.8571) (458, 0.8571) (459, 0.8571) (460, 0.8571) (461, 0.8571) (462, 0.8571) (463, 0.8571) (464, 0.8571) (465, 0.8571) (466, 0.8571) (467, 0.8571) (468, 0.8571) (469, 0.8571) (470, 0.8571) (471, 0.8571) (472, 0.8571) (473, 0.8571) (474, 0.8571) (475, 0.8571) (476, 0.8571) (477, 0.8571) (478, 0.8571) (479, 0.8571) (480, 0.8571) (481, 0.8571) (482, 0.8571) (483, 0.8528) (484, 0.8528) (485, 0.8528) (486, 0.8528) (487, 0.8528) (488, 0.8571) (489, 0.8571) (490, 0.8528) (491, 0.8528) (492, 0.8528) (493, 0.8528) (494, 0.8528) (495, 0.8528) (496, 0.8571) (497, 0.8571) (498, 0.8528) (499, 0.8528) (500, 0.8528) };
        \end{groupplot}
        \end{tikzpicture}
        \caption{
            PA for varying dataset sizes at a fine-grained scale.
        } 
        \label{fig:size-ablation}
    \end{subfigure}
    \hfill
    \begin{subfigure}[b]{0.48\textwidth}
        \centering
        \pgfplotsset{
            footnotesize,
            samples=10,
            xmin=0.5, xmax=3.5,
            ymin=0.75, ymax=0.87,
            xtick={1,2,3},
            xticklabels={100,250,500},
            ytick={0.76,0.78,0.80,0.82,0.84,0.86},
            yticklabels={0.76,0.78,0.80,0.82,0.84,0.86},
            grid style={dashed},
            grid=both,
            xlabel=Number of instances in test set,
            x label style={at={(axis description cs:0.5,-0.15)},anchor=north},
        }
        
        \begin{tikzpicture}
        \begin{groupplot}[
            group style = {group size = 1 by 1, horizontal sep = 24pt},
            width = \textwidth, 
            height = 4cm
        ]
            \nextgroupplot[
                align = center,
                legend style={
                    at={(0.5,1.25)},
                    anchor=north,
                    legend columns=2,
                    column sep=10pt,
                    draw=none,
                },
                legend cell align={left},
                axis x line*=bottom,
                axis y line*=left,
                y label style={at={(axis description cs:-0.15,0.485)},anchor=south},
                xtick pos=bottom,
                ytick pos=left,
            ]
            \addplot[color=CustomBlue!40, line width=1.1pt, mark=square*, mark size=2.5pt] coordinates {(1, 0.7619)};
            \addplot[color=CustomBlue!40, line width=1.1pt, mark=triangle*, mark size=2.5pt] coordinates {(1, 0.7792)};
            \addplot[color=CustomBlue!40, line width=1.1pt, mark=diamond*, mark size=2.5pt] coordinates {(1, 0.8312)};
            \addplot[color=CustomBlue!40, line width=1.1pt, mark=*, mark size=2.5pt] coordinates {(1, 0.8355)};

            \addplot[color=CustomBlue!60, line width=1.1pt, mark=square*, mark size=2.5pt] coordinates {(2, 0.7749)};
            \addplot[color=CustomBlue!60, line width=1.1pt, mark=triangle*, mark size=2.5pt] coordinates {(2, 0.7965)};
            \addplot[color=CustomBlue!60, line width=1.1pt, mark=diamond*, mark size=2.5pt] coordinates {(2, 0.8355)};
            \addplot[color=CustomBlue!60, line width=1.1pt, mark=*, mark size=2.5pt] coordinates {(2, 0.8485)};

            \addplot[color=CustomBlue, line width=1.1pt, mark=square*, mark size=2.5pt] coordinates {(3, 0.7749)};
            \addplot[color=CustomBlue, line width=1.1pt, mark=triangle*, mark size=2.5pt] coordinates {(3, 0.8009)};
            \addplot[color=CustomBlue, line width=1.1pt, mark=diamond*, mark size=2.5pt, draw=black, on above layer] coordinates {(3, 0.8571)};
            \addplot[color=CustomBlue, line width=1.1pt, mark=*, mark size=2.5pt] coordinates {(3, 0.8615)};

            \addplot[black!50, line width=0.75pt, dashed, forget plot] coordinates {(1, 0.7619) (2, 0.7749) (3, 0.7749)};
            \addplot[black!50, line width=0.75pt, dashed, forget plot] coordinates {(1, 0.7792) (2, 0.7965) (3, 0.8009)};
            \addplot[black!50, line width=0.75pt, dashed, forget plot] coordinates {(1, 0.8312) (2, 0.8355) (3, 0.8571)};
            \addplot[black!50, line width=0.75pt, dashed, forget plot] coordinates {(1, 0.8355) (2, 0.8485) (3, 0.8615)};

            \node[align=center, anchor=north, yshift=-1pt, text width=4cm, font=\scriptsize] at (axis cs:2, 0.7725) {Topic only};
            \node[align=center, anchor=north, yshift=-1pt, text width=4cm, font=\scriptsize] at (axis cs:1.25, 0.81) {+ subtopic};
            \node[align=center, anchor=north, yshift=-1pt, text width=4cm, font=\scriptsize] at (axis cs:1.25, 0.83) {+ other variables};
            \node[align=center, anchor=north, yshift=-1pt, text width=4cm, font=\scriptsize] at (axis cs:1.25, 0.867) {+ reference};
            
            \node[align=center, anchor=north, yshift=-1pt, text width=4cm, font=\scriptsize] at (axis cs:2.8, 0.847) {Default \\ configuration};
            \draw[<-, bend left=45, line width=0.8pt] 
            (axis cs:2.8, 0.84) to[out=0,in=0] (axis cs:2.98, 0.851);

        \end{groupplot}
        \end{tikzpicture}
        \caption{
            PA for varying dataset sizes (hue) and meta prompt setups (marker).
        } 
        \label{fig:variety-ablations}
    \end{subfigure}
    \caption{Analysis of factors affecting the PA of the ZSB for general capabilities in English.}
    \label{fig:combined-ablations}
\end{figure}

\paragraph{Dataset variety is a strong driver of benchmark performance.}
Figure~\ref{fig:variety-ablations} shows the performance of different iterations of the data generation process, where we vary the dataset size, the number of attributes used in the meta prompt, and the usage of a reference.
Adding a reference leads to marginal improvements in performance.
Crucially, adding more attribute variety in the meta prompt can lead to improvements of more than 6 PA points.
This finding suggests that performance can be further increased with more variety through adding more attributes, or even leveraging external sources of data.

\section{Related Work}\label{sec:rw}
\paragraph{Fully-automated evaluation suites.} We propose and show the effectiveness of a simple framework to automate test data creation and evaluation for language modeling tasks.
Previous work is mainly inspired by the Chatbot Arena~\citep{zheng2023judging}, where both steps are crowd-sourced.
While involving humans enables the collection of gold standard test data and evaluations, it has two shortcomings: 1) evaluation is slow and public, so it is unsuitable for model development; 2) it requires a critical mass of users to be effective, which limits its reach to less popular, yet relevant, tasks.
\citet{li2024crowdsourced} propose a method for creating synthetic test sets for chatbots by leveraging pre-existing data from the Chatbot Arena. 
While this solves the first issue, it does not solve the second, since it requires pre-existing data.
\citet{zhao2024auto}~propose automating both data creation and evaluation, but they focus solely on chatbot capabilities in English and Chinese,\footnote{They do not compare against other baselines for Chinese, and their reported performance for the Open LLM Leaderboard is low (0.325 Spearman), raising questions about their meta-evaluation setup.} leaving out other real-world use-cases and tasks.
\citet{hu2025dual} also focus solely on dialogue response generation and summarization in English.
\citet{luo2024videoautoarena}~extend the approach to the video modality and perform in-house meta-evaluation, since there is no established gold ranking.
\citet{butt2024benchagents} propose a framework to create benchmarks with several LLMs as agents, and test seven LLMs on generated calendar scheduling and constrained text generation benchmarks (both in English).
ZSB builds on these works by facilitating the exploration of a wider variety of languages and tasks, including multimodal ones.
Furthermore, we dissect the impact of different components of our framework on benchmark performance.

\paragraph{Automatic evaluation with static test sets.} Automatic evaluation is an integral part of language model development.
The most common approach is to create a human-annotated test set, and then evaluate the model on it using some automatic metric.
Metrics can be intrinsic, like accuracy for multiple-choice tests like MMLU~\citep{hendrycks2measuring}, and \bleu{}~\citep{papineni2002bleu} for MT, or learned models themselves, like \comet{}~\citep{rei2020comet} for MT, or \bleurt{}~\citep{sellam2020bleurt} for text generation.
The judgments of the latter are usually more costly to obtain, but correlate better with human judgments.
With the rapid expansion of capabilities in language models, the development of new task-specific metrics has become harder.
Thus, LLMs have been increasingly used as judges~\citep[LLM-as-a-judge]{zheng2023judging,gu2024survey,li2024generation,li2024llms}.
Some judges, like Prometheus~\citep{kim2023prometheus}, are explicitly designed for this purpose through finetuning, while others are simply general-purpose models~\citep{zheng2023judging,kocmi2023gemba}.
We employ judge LLMs in our framework, where they excel across all tasks.

\paragraph{Synthetic test sets.} Another consequence of the increasing capabilities of language models is that test sets saturate faster.
To address this challenge, several work introduce synthetically-generated test sets~\citep{wu2024unigen,weston2015towards,gandhi2024understanding,le2019revisiting,clark2020transformers,saparov2022language,sinha2019clutrr,dalvi2021explaining,sprague2022natural,kazemi2024boardgameqa}.
The advantage of this approach is that it is much more scalable than human-annotated test sets, and they can be enhanced easily as language models improve.
One notable is example is MuSR~\citep{spraguemusr}, a multiple-choice reasoning benchmark used in the widely-adopted Open LLM Leaderboard.
Our framework benefits from the main advantages of synthetic test sets: scalability, and the capacity to create increasingly varied and challenging benchmarks as models improve.

\section{Conclusion}\label{sec:conc}
We presented Zero-shot Benchmarking (ZSB), a framework for creating high-quality benchmarks for language modeling tasks by automating both data creation and evaluation with language models. 
Through extensive experiments across five text tasks and a multi-modal one, we showed that our framework produces system rankings that correlate well with human judgments, outperforming widely-used standard benchmarks. 
Our ablation studies revealed several key findings: 1) open models can be used to create high-quality benchmarks, though judge model size is crucial; 2) direct assessments are a viable alternative to pairwise comparisons; 3) dataset variety is more important than size for benchmark quality.

The implications of our work are significant. 
ZSB enables rapid development of benchmarks for specialized tasks where collecting human-annotated data would be costly or impractical.
Furthermore, its flexibility allows benchmarks to evolve alongside improvements in language model capabilities, ensuring continued relevance. 
We release all our benchmarks and code to facilitate adoption and further research in this direction.
In the future we would like to explore applying the framework to other modalities and tasks, investigating methods for synthetic image generation in multimodal benchmarks, and developing techniques to further increase dataset variety (e.g., through the retrieval of relevant context to the task). 
Additionally, studying the relationship between benchmark performance and real-world utility remains an important open question.

\section*{Acknowledgements} 
We thank Sweta Agrawal, Giuseppe Attanasio, Duarte Alves, António Farinhas, Haau-Sing Li, and Beatriz Canaverde for their constructive feedback on the paper. We acknowledge EuroHPC JU for awarding the project ID EHPC-AI-2024A01-085 access to MareNostrum 5 ACC. This work was supported by EU’s Horizon Europe Research and Innovation Actions (UTTER, contract 101070631), by the project DECOLLAGE (ERC-2022-CoG 101088763), by the Portuguese Recovery and Resilience Plan through project C64500888200000055 (Center for Responsible AI), and by Fundação para a Ciência e Tecnologia through contract UIDB/50008/2020.

\section*{Reproducibility Statement} We release all our benchmarks and code to reproduce our experiments and to create new benchmarks.
Part of our experiments rely on closed models, which may become unavailable in the future, posing a potential challenge for reproducibility.

\section*{Ethics Statement}
Our work focuses on automating benchmark creation, which raises two key concerns. 
First, synthetic data may encode biases present in the models used to create it. 
While we include safety scores for transparency, these are not a complete solution---users should carefully review generated instances before deployment. 
Second, automated evaluation could be misused to claim superiority without proper validation. 
We emphasize that our benchmarks should complement, not replace, careful human evaluation and real-world testing.
We release our code and data to enable scrutiny and improvement by the research community.

\bibliography{custom,anthology}
\bibliographystyle{colm2025_conference}

\appendix
\section{Prompts and ZSB Examples}\label{sec:apx-prompts-and-examples}

\subsection{ZSB Meta-prompts for Data Generation}\label{subsec:apx-meta-prompts}

\begin{figure}[H]
    \small
    \renewcommand\tabularxcolumn[1]{m{#1}}
   \centering
\begin{tabularx}{\linewidth}{*{1}{>{\arraybackslash}X}}
You are tasked with creating a diverse and engaging prompt for a chatbot arena. This prompt will be used to test and compare the capabilities of different language models. Your goal is to generate a question or prompt that will challenge these models and showcase their strengths or weaknesses. Also, generate a reference answer to your prompt that will serve as a benchmark for evaluating the models' responses. \\
\hfill \\
Here are the input variables you will use to craft your prompt: \\
- Language: \$\{language\} \\
- Topic: \$\{topic\} \\
- Subtopic: \$\{subtopic\} \\
- Difficulty: \$\{difficulty\} \\
- Style: \$\{style\} \\
- Writer: \$\{writer\} \\
- Writing proficiency: \$\{writing\_proficiency\} \\
- Prompt length: \$\{length\} \\
\hfill \\
Guidelines for creating the prompt: \\
\hfill \\
1. Abide strictly by the input variables provided. \\
\hfill \\
2. Ensure that your prompt is open-ended enough to allow for varied and interesting responses from different language models. \\
\hfill \\
3. Avoid prompts that are overly specific to a particular AI model's capabilities or training data. \\
\hfill \\
4. Create a prompt that is engaging and thought-provoking, encouraging creative or analytical thinking. \\
\hfill \\
Generate output in the following format: \\
\hfill \\
$<$START OF PROMPT$>$ \\
$[$Your generated prompt here. IMPORTANT: include only the prompt.$]$ \\
$<$END OF PROMPT$>$ \\
\hfill \\
$<$START OF REFERENCE$>$ \\
$[$A reference answer to your prompt. IMPORTANT: include only the reference answer.$]$ \\
$<$END OF REFERENCE$>$ \\
\hfill \\
Ensure the generated prompt is in the requested language. Remember to abide strictly by the provided input variables and the requested format.
\end{tabularx}
\caption{Meta-prompt for multilingual general capabilities.}\label{fig:meta_prompt_general_capabilities}
\end{figure}

\begin{figure}[H]
    \small
    \renewcommand\tabularxcolumn[1]{m{#1}}
   \centering
\begin{tabularx}{\linewidth}{*{1}{>{\arraybackslash}X}}
You are a multilingual content creator and translation expert. Your task is to generate a comprehensive translation exercise based on the given attributes. Follow these instructions carefully: \\
\hfill \\
1. Review the following input variables: \\
- Source language: \$\{source\_language\} \\
- Target language: \$\{target\_language\} \\
- Topic: \$\{topic\} \\
- Subtopic: \$\{subtopic\} \\
- Source Length: \$\{source\_length\} \\
- Style: \$\{style\} \\
\hfill \\
2. Generate a source text:
Create an original text in the source language, adhering to the specified topic, subtopic, and length. The text should be coherent, informative, and suitable for translation. \\
\hfill \\
3. Generate a reference translation:
Produce a high-quality, fluent translation of the source text in the target language. This translation should serve as a reference for evaluating other translations. IT IS CRUCIAL THAT THE REFERENCE TRANSLATION SOUNDS NATURAL IN THE TARGET LANGUAGE. \\
\hfill \\
Format your output as follows: \\
\hfill \\
$<$START OF SOURCE$>$ \\
$[$INSERT THE SOURCE TEXT HERE$]$ \\
$<$END OF SOURCE$>$ \\
\hfill \\
$<$START OF REFERENCE TRANSLATION$>$ \\
$[$INSERT THE REFERENCE TRANSLATION HERE$]$ \\
$<$END OF REFERENCE TRANSLATION$>$ \\
\hfill \\
Ensure that your response is comprehensive, coherent, and follows all the instructions provided above. Abide strictly by the requested format and generated until the end of the requested output.
\end{tabularx}
\caption{ZSB meta-prompt for translation.}\label{fig:meta_prompt_translation}
\end{figure}

\begin{figure}[H]
    \small
    \renewcommand\tabularxcolumn[1]{m{#1}}
   \centering
\begin{tabularx}{\linewidth}{*{1}{>{\arraybackslash}X}}
You are tasked with creating a diverse and engaging prompt for a vision language model chatbot arena, given an input image. This prompt will be used to test and compare the capabilities of different vision language models when they analyze images. Your goal is to generate a question or prompt that will challenge these models and showcase their strengths or weaknesses in image understanding, analysis, and reasoning. Also, generate a reference answer that will serve as a benchmark for evaluating the models' responses. \\
\hfill \\
Guidelines for creating the prompt:
- Ensure the prompt is relevant to the attached image. \\
- Create a prompt that requires both visual analysis and reasoning capabilities. \\
- Do not describe the image in the prompt; let the model analyze it. \\
\hfill \\
Ensure your prompt encourages the model to: \\
- Describe specific details in the image. \\
- Make connections between different elements. \\
- Draw conclusions or insights. \\
- Provide explanations or interpretations. \\
- Consider context and implications. \\
\hfill \\  
\hfill \\  
Avoid prompts that: \\
- Can be answered without looking at the image. \\
- Focus only on simple object detection. \\
- Are biased toward specific vision models. \\
\hfill \\
\hfill \\
Make prompts that test various vision-language capabilities such as: \\
- Spatial reasoning \\
- Object relationships \\
- Visual detail recognition \\
- Context understanding \\
- Abstract concept interpretation \\
- Temporal reasoning from static images \\
- Cultural or symbolic understanding \\
\hfill \\
\hfill \\
\hfill \\
Generate output in the following format:
\hfill \\
$<$START OF PROMPT$>$ \\
$[$Your generated prompt here. This should be a question or instruction that will be paired with the image for the model to analyze.$]$ \\
$<$END OF PROMPT$>$ \\
\hfill \\
$<$START OF REFERENCE$>$ \\
$[$A reference answer, and only the answer, to the generated prompt.$]$ \\
$<$END OF REFERENCE$>$ \\
\hfill \\
Remember to abide strictly by the provided input variables and the requested format.
\end{tabularx}
\caption{ZSB meta-prompt for VLM general capabilities. For each generated instance, this prompt is accompanied with some image.}\label{fig:meta_prompt_vlm_general_capabilities}
\end{figure}

\subsection{Meta-prompt Attributes and Distributions}\label{apx-attrs-dists}

\begin{table}[H]
    \centering
    \setlength{\tabcolsep}{3pt}
\renewcommand{\arraystretch}{1.3}
\scriptsize
\begin{tabular}{l c c}
    \toprule
    \textbf{Meta-prompt Attribute} & \textbf{LLM General Capabilities} & \textbf{Translation} \\
    \midrule			
        \multicolumn{3}{l}{\textbf{Topic \& Subtopic}} \\
        Tech innovation & 1.60\% & 1.80\% \\
            \quad Artificial intelligence & 0.00\% & 0.20\% \\
            \quad Quantum computing & 0.40\% & 0.20\% \\
            \quad Robotics & 0.40\% & 0.00\% \\
            \quad 5g/6g networks & 0.40\% & 0.00\% \\
            \quad Biotechnology & 0.00\% & 0.40\% \\
            \quad Green tech & 0.00\% & 0.40\% \\
            \quad Edge computing & 0.20\% & 0.20\% \\
            \quad Cybersecurity & 0.20\% & 0.40\% \\
        Global markets & 1.80\% & 2.00\% \\
            \quad Stock exchanges & 0.00\% & 0.20\% \\
            \quad Cryptocurrency & 0.20\% & 0.60\% \\
            \quad International trade & 0.60\% & 0.20\% \\
            \quad Foreign investment & 0.20\% & 0.00\% \\
            \quad Commodity markets & 0.20\% & 0.40\% \\
            \quad Emerging markets & 0.40\% & 0.20\% \\
            \quad Foreign exchange & 0.00\% & 0.40\% \\
            \quad Market regulations & 0.60\% & 0.00\% \\
        Environmental policy & 1.20\% & 1.00\% \\
            \quad Carbon trading & 0.20\% & 0.20\% \\
            \quad Renewable energy initiatives & 0.00\% & 0.40\% \\
            \quad Wildlife protection & 0.60\% & 0.20\% \\
            \quad Urban planning & 1.00\% & 1.20\% \\
            \quad Waste management & 0.20\% & 0.00\% \\
            \quad Climate agreements & 0.00\% & 0.20\% \\
            \quad Marine conservation & 0.00\% & 0.00\% \\
        Public health & 1.40\% & 1.80\% \\
            \quad Disease prevention & 0.40\% & 0.00\% \\
            \quad Healthcare systems & 0.40\% & 0.20\% \\
            \quad Vaccination programs & 0.00\% & 0.00\% \\
            \quad Mental health services & 0.40\% & 0.80\% \\
            \quad Maternal health & 0.00\% & 0.20\% \\
            \quad Epidemiology & 0.20\% & 0.40\% \\
            \quad Health technology & 0.00\% & 0.20\% \\
        Urban development & 1.40\% & 1.00\% \\
            \quad Smart cities & 0.20\% & 0.20\% \\
            \quad Public transportation & 0.40\% & 0.00\% \\
            \quad Housing projects & 0.60\% & 0.00\% \\
            \quad Green spaces & 0.00\% & 0.40\% \\
            \quad Infrastructure & 0.20\% & 0.40\% \\
            \quad Urban planning & 1.00\% & 1.20\% \\
            \quad Sustainable development & 0.40\% & 0.20\% \\
    \bottomrule
\end{tabular}
    \caption{
        Part 1 of all meta-prompt attribute values and their prevalence on all Zero-Shots Benchmarks released.
        Distributions are the same across languages and language pairs, respectively.
        The VLM task meta-prompt does not have attributes, only an image.
    }
    \label{tab:attrs-and-dists-1}
\end{table}

\begin{table}[H]
    \centering
    \setlength{\tabcolsep}{3pt}
\renewcommand{\arraystretch}{1.3}
\scriptsize
\begin{tabular}{l c c}
    \toprule
    \textbf{Meta-prompt Attribute} & \textbf{LLM General Capabilities} & \textbf{Translation} \\
    \midrule			
        \multicolumn{3}{l}{\textbf{Topic}} \\
        International relations & 1.80\% & 2.00\% \\
            \quad Diplomatic missions & 0.20\% & 0.00\% \\
            \quad Trade agreements & 0.40\% & 0.60\% \\
            \quad Cultural exchange & 0.00\% & 0.20\% \\
            \quad Peace negotiations & 0.20\% & 0.00\% \\
            \quad International aid & 0.00\% & 0.40\% \\
            \quad Global security & 0.60\% & 0.40\% \\
            \quad Regional alliances & 0.40\% & 0.40\% \\
        Education reform & 1.40\% & 1.20\% \\
            \quad Digital learning & 0.40\% & 0.20\% \\
            \quad Curriculum changes & 0.20\% & 0.00\% \\
            \quad Teacher training & 0.20\% & 0.40\% \\
            \quad Assessment methods & 0.00\% & 0.40\% \\
            \quad Special education & 0.20\% & 0.20\% \\
            \quad Higher education & 0.20\% & 0.00\% \\
            \quad Stem initiatives & 0.20\% & 0.00\% \\
        Cultural trends & 1.20\% & 2.20\% \\
            \quad Social media influence & 0.00\% & 0.40\% \\
            \quad Fashion movements & 0.00\% & 0.20\% \\
            \quad Entertainment trends & 0.00\% & 0.40\% \\
            \quad Digital culture & 0.40\% & 0.20\% \\
            \quad Cultural festivals & 0.40\% & 1.00\% \\
            \quad Art movements & 0.20\% & 0.20\% \\
            \quad Pop culture & 1.40\% & 0.60\% \\
        Scientific discoveries & 1.60\% & 1.00\% \\
            \quad Space exploration & 0.20\% & 0.00\% \\
            \quad Medical breakthroughs & 0.00\% & 0.20\% \\
            \quad Physics advances & 0.20\% & 0.00\% \\
            \quad Genetic research & 0.20\% & 0.00\% \\
            \quad Archaeological finds & 0.60\% & 0.20\% \\
            \quad Marine biology & 0.20\% & 0.00\% \\
            \quad Climate science & 0.40\% & 0.80\% \\
        Economic policy & 1.40\% & 1.00\% \\
            \quad Monetary policy & 0.40\% & 0.00\% \\
            \quad Fiscal measures & 0.40\% & 0.20\% \\
            \quad Trade regulations & 0.20\% & 0.20\% \\
            \quad Tax reform & 0.00\% & 0.00\% \\
            \quad Employment policy & 0.20\% & 0.20\% \\
            \quad Banking regulations & 0.00\% & 0.00\% \\
            \quad Economic stimulus & 0.20\% & 0.40\% \\
        Sports industry & 0.80\% & 1.40\% \\
            \quad Professional leagues & 0.00\% & 0.00\% \\
            \quad Sports technology & 0.20\% & 0.20\% \\
            \quad Athletic training & 0.20\% & 0.20\% \\
            \quad Sports medicine & 0.00\% & 0.60\% \\
            \quad E-sports & 0.60\% & 0.20\% \\
            \quad Sports management & 0.20\% & 0.40\% \\
            \quad Athletic equipment & 0.00\% & 0.00\% \\
        Media \& entertainment & 1.80\% & 1.80\% \\
            \quad Streaming services & 0.40\% & 0.00\% \\
            \quad Film production & 0.20\% & 0.20\% \\
            \quad Gaming industry & 0.40\% & 0.00\% \\
            \quad Digital media & 0.20\% & 0.20\% \\
            \quad Publishing & 0.20\% & 0.60\% \\
            \quad Broadcasting & 0.20\% & 0.60\% \\
            \quad Content creation & 0.40\% & 0.40\% \\
    \bottomrule
\end{tabular}
    \caption{
        Part 2 of all meta-prompt attribute values and their prevalence on all Zero-Shots Benchmarks released.
        Distributions are the same across languages and language pairs, respectively.
        The VLM task meta-prompt does not have attributes, only an image.
    }
    \label{tab:attrs-and-dists-2}
\end{table}

\begin{table}[H]
    \centering
    \setlength{\tabcolsep}{3pt}
\renewcommand{\arraystretch}{1.3}
\scriptsize
\begin{tabular}{l c c}
    \toprule
    \textbf{Meta-prompt Attribute} & \textbf{LLM General Capabilities} & \textbf{Translation} \\
    \midrule			
        \multicolumn{3}{l}{\textbf{Topic}} \\
        Workplace transformation & 1.40\% & 0.60\% \\
            \quad Remote work & 0.00\% & 0.00\% \\
            \quad Office technology & 0.60\% & 0.00\% \\
            \quad Employee wellness & 0.20\% & 0.00\% \\
            \quad Corporate culture & 0.00\% & 0.20\% \\
            \quad Hr innovation & 0.20\% & 0.00\% \\
            \quad Workplace safety & 0.20\% & 0.20\% \\
            \quad Professional development & 0.20\% & 0.20\% \\
        Transportation \& mobility & 0.40\% & 1.40\% \\
            \quad Electric vehicles & 0.20\% & 0.40\% \\
            \quad Autonomous driving & 0.00\% & 0.20\% \\
            \quad Public transit & 0.00\% & 0.40\% \\
            \quad Aviation & 0.20\% & 0.20\% \\
            \quad Ride sharing & 0.00\% & 0.20\% \\
            \quad Maritime transport & 0.00\% & 0.20\% \\
            \quad Urban mobility & 0.20\% & 0.20\% \\
        Food \& agriculture & 1.60\% & 0.80\% \\
            \quad Sustainable farming & 0.40\% & 0.00\% \\
            \quad Food technology & 0.20\% & 0.40\% \\
            \quad Organic production & 0.40\% & 0.40\% \\
            \quad Agricultural policy & 0.00\% & 0.00\% \\
            \quad Food safety & 0.00\% & 0.00\% \\
            \quad Urban farming & 0.00\% & 0.20\% \\
            \quad Agricultural trade & 0.60\% & 0.00\% \\
        Medical \& healthcare & 0.80\% & 1.00\% \\
            \quad Telemedicine & 0.20\% & 0.00\% \\
            \quad Medical devices & 0.20\% & 0.20\% \\
            \quad Pharmaceutical research & 0.00\% & 0.00\% \\
            \quad Healthcare it & 0.20\% & 0.00\% \\
            \quad Patient care & 0.00\% & 0.40\% \\
            \quad Medical insurance & 0.20\% & 0.40\% \\
            \quad Clinical trials & 0.40\% & 0.00\% \\
        Legal \& compliance & 1.00\% & 1.40\% \\
            \quad Corporate law & 0.00\% & 0.20\% \\
            \quad Intellectual property & 0.00\% & 0.20\% \\
            \quad Data protection & 0.20\% & 0.20\% \\
            \quad Regulatory compliance & 0.00\% & 0.60\% \\
            \quad International law & 0.20\% & 0.00\% \\
            \quad Consumer rights & 0.00\% & 0.00\% \\
            \quad Legal technology & 0.60\% & 0.40\% \\
        E-commerce \& retail & 1.40\% & 2.00\% \\
            \quad Online marketplaces & 0.20\% & 0.20\% \\
            \quad Digital payment & 0.40\% & 0.20\% \\
            \quad Retail technology & 0.40\% & 0.00\% \\
            \quad Supply chain & 0.00\% & 0.60\% \\
            \quad Customer experience & 0.00\% & 0.00\% \\
            \quad Mobile commerce & 0.00\% & 0.40\% \\
            \quad Retail analytics & 0.40\% & 0.60\% \\
        Financial services & 1.80\% & 1.20\% \\
            \quad Digital banking & 0.80\% & 0.00\% \\
            \quad Investment management & 0.20\% & 0.20\% \\
            \quad Insurance & 0.40\% & 0.20\% \\
            \quad Payment systems & 0.00\% & 0.40\% \\
            \quad Wealth management & 0.20\% & 0.40\% \\
            \quad Risk management & 0.20\% & 0.00\% \\
            \quad Financial technology & 0.00\% & 0.00\% \\
        \bottomrule
\end{tabular}
    \caption{
        Part 3 of all meta-prompt attribute values and their prevalence on all Zero-Shots Benchmarks released.
        Distributions are the same across languages and language pairs, respectively.
        The VLM task meta-prompt does not have attributes, only an image.
    }
    \label{tab:attrs-and-dists-3}
\end{table}

\begin{table}[H]
    \centering
    \setlength{\tabcolsep}{3pt}
\renewcommand{\arraystretch}{1.3}
\scriptsize
\begin{tabular}{l c c}
    \toprule
    \textbf{Meta-prompt Attribute} & \textbf{LLM General Capabilities} & \textbf{Translation} \\
    \midrule			
        \multicolumn{3}{l}{\textbf{Topic}} \\
        Gaming \& software & 1.40\% & 0.80\% \\
            \quad Game development & 0.40\% & 0.00\% \\
            \quad Cloud gaming & 0.60\% & 0.00\% \\
            \quad Mobile games & 0.40\% & 0.00\% \\
            \quad Gaming hardware & 0.20\% & 0.00\% \\
            \quad Software development & 0.00\% & 0.40\% \\
            \quad Virtual reality & 0.00\% & 0.20\% \\
            \quad Gaming communities & 0.00\% & 0.20\% \\
        Marketing \& advertising & 0.60\% & 0.60\% \\
            \quad Digital marketing & 0.20\% & 0.20\% \\
            \quad Brand strategy & 0.00\% & 0.20\% \\
            \quad Social media marketing & 0.20\% & 0.20\% \\
            \quad Content marketing & 0.20\% & 0.00\% \\
            \quad Market research & 0.00\% & 0.00\% \\
            \quad Advertising technology & 0.00\% & 0.00\% \\
            \quad Campaign management & 0.00\% & 0.00\% \\
        Government documentation & 0.80\% & 1.40\% \\
            \quad Policy documents & 0.00\% & 0.40\% \\
            \quad Legal records & 0.00\% & 0.20\% \\
            \quad Public notices & 0.20\% & 0.00\% \\
            \quad Official forms & 0.20\% & 0.40\% \\
            \quad Legislative documents & 0.20\% & 0.00\% \\
            \quad Regulatory guidelines & 0.00\% & 0.20\% \\
            \quad Administrative procedures & 0.20\% & 0.20\% \\
        Academic research & 1.40\% & 0.40\% \\
            \quad Scientific publications & 0.80\% & 0.20\% \\
            \quad Research methodology & 0.20\% & 0.00\% \\
            \quad Peer review & 0.00\% & 0.00\% \\
            \quad Academic collaboration & 0.20\% & 0.00\% \\
            \quad Research funding & 0.00\% & 0.20\% \\
            \quad Data analysis & 0.20\% & 0.00\% \\
            \quad Research ethics & 0.00\% & 0.00\% \\
        Patents \& intellectual property & 0.60\% & 1.00\% \\
            \quad Patent applications & 0.00\% & 0.20\% \\
            \quad Trademark registration & 0.00\% & 0.00\% \\
            \quad Copyright protection & 0.00\% & 0.40\% \\
            \quad Ip litigation & 0.20\% & 0.00\% \\
            \quad International patents & 0.20\% & 0.00\% \\
            \quad Trade secrets & 0.00\% & 0.20\% \\
            \quad Ip strategy & 0.20\% & 0.20\% \\
        Manufacturing \& safety & 1.00\% & 1.20\% \\
            \quad Quality control & 0.20\% & 0.20\% \\
            \quad Industrial safety & 0.20\% & 0.40\% \\
            \quad Production processes & 0.20\% & 0.20\% \\
            \quad Workplace standards & 0.00\% & 0.40\% \\
            \quad Equipment safety & 0.20\% & 0.00\% \\
            \quad Safety regulations & 0.20\% & 0.00\% \\
            \quad Risk assessment & 0.20\% & 0.20\% \\
        Tourism \& hospitality & 0.80\% & 1.20\% \\
            \quad Hotel management & 0.00\% & 0.20\% \\
            \quad Travel services & 0.00\% & 0.20\% \\
            \quad Tourism marketing & 0.20\% & 0.20\% \\
            \quad Cultural tourism & 0.00\% & 0.00\% \\
            \quad Eco-tourism & 0.20\% & 0.40\% \\
            \quad Event planning & 0.20\% & 0.00\% \\
            \quad Customer service & 0.20\% & 0.20\% \\
    \bottomrule
\end{tabular}
    \caption{
        Part 4 of all meta-prompt attribute values and their prevalence on all Zero-Shots Benchmarks released.
        Distributions are the same across languages and language pairs, respectively.
        The VLM task meta-prompt does not have attributes, only an image.
    }
    \label{tab:attrs-and-dists-4}
\end{table}

\begin{table}[H]
    \centering
    \setlength{\tabcolsep}{3pt}
\renewcommand{\arraystretch}{1.3}
\scriptsize
\begin{tabular}{l c c}
    \toprule
    \textbf{Meta-prompt Attribute} & \textbf{LLM General Capabilities} & \textbf{Translation} \\
    \midrule			
        \multicolumn{3}{l}{\textbf{Topic}} \\
        Religious \& cultural studies & 1.40\% & 1.00\% \\
            \quad Religious traditions & 0.00\% & 0.20\% \\
            \quad Cultural heritage & 1.60\% & 3.20\% \\
            \quad Interfaith dialogue & 0.40\% & 0.40\% \\
            \quad Religious practices & 0.00\% & 0.20\% \\
            \quad Cultural anthropology & 0.60\% & 0.20\% \\
            \quad Sacred texts & 0.40\% & 0.00\% \\
            \quad Religious education & 0.00\% & 0.00\% \\
        Insurance \& risk management & 1.20\% & 0.80\% \\
            \quad Insurance products & 0.40\% & 0.20\% \\
            \quad Risk assessment & 0.20\% & 0.20\% \\
            \quad Claims processing & 0.00\% & 0.20\% \\
            \quad Underwriting & 0.20\% & 0.00\% \\
            \quad Insurance technology & 0.20\% & 0.00\% \\
            \quad Regulatory compliance & 0.00\% & 0.60\% \\
            \quad Risk mitigation & 0.20\% & 0.00\% \\
        Consumer electronics & 1.00\% & 1.20\% \\
            \quad Mobile devices & 0.00\% & 0.20\% \\
            \quad Smart home & 0.40\% & 0.40\% \\
            \quad Wearable technology & 0.00\% & 0.40\% \\
            \quad Audio equipment & 0.20\% & 0.20\% \\
            \quad Display technology & 0.20\% & 0.00\% \\
            \quad Gaming hardware & 0.20\% & 0.00\% \\
            \quad Personal computing & 0.00\% & 0.20\% \\
        Pharmaceutical industry & 0.80\% & 0.80\% \\
            \quad Drug development & 0.00\% & 0.20\% \\
            \quad Clinical trials & 0.40\% & 0.00\% \\
            \quad Regulatory approval & 0.00\% & 0.00\% \\
            \quad Manufacturing & 0.20\% & 0.40\% \\
            \quad Research \& development & 0.20\% & 0.20\% \\
            \quad Market access & 0.00\% & 0.00\% \\
            \quad Patient safety & 0.00\% & 0.00\% \\
        Fashion \& apparel & 1.40\% & 1.00\% \\
            \quad Fashion design & 0.20\% & 0.00\% \\
            \quad Textile industry & 0.20\% & 0.20\% \\
            \quad Sustainable fashion & 0.20\% & 0.40\% \\
            \quad Retail fashion & 0.00\% & 0.00\% \\
            \quad Fashion technology & 0.00\% & 0.20\% \\
            \quad Luxury brands & 0.80\% & 0.80\% \\
            \quad Fashion marketing & 0.20\% & 0.00\% \\
        Beauty \& cosmetics & 0.80\% & 1.20\% \\
            \quad Skincare & 0.20\% & 0.00\% \\
            \quad Makeup products & 0.20\% & 0.00\% \\
            \quad Natural cosmetics & 0.00\% & 0.40\% \\
            \quad Beauty technology & 0.40\% & 0.20\% \\
            \quad Product development & 0.00\% & 0.40\% \\
            \quad Marketing & 0.00\% & 0.20\% \\
            \quad Sustainability & 0.20\% & 0.00\% \\
        Home \& living & 1.00\% & 0.60\% \\
            \quad Interior design & 0.00\% & 0.60\% \\
            \quad Smart home & 0.40\% & 0.40\% \\
            \quad Furniture & 0.40\% & 0.00\% \\
            \quad Home improvement & 0.20\% & 0.20\% \\
            \quad Sustainable living & 0.00\% & 0.00\% \\
            \quad Home technology & 0.20\% & 0.00\% \\
            \quad Decorative arts & 0.20\% & 0.20\% \\
    \bottomrule
\end{tabular}
    \caption{
        Part 5 of all meta-prompt attribute values and their prevalence on all Zero-Shots Benchmarks released.
        Distributions are the same across languages and language pairs, respectively.
        The VLM task meta-prompt does not have attributes, only an image.
    }
    \label{tab:attrs-and-dists-5}
\end{table}

\begin{table}[H]
    \centering
    \setlength{\tabcolsep}{3pt}
\renewcommand{\arraystretch}{1.3}
\scriptsize
\begin{tabular}{l c c}
    \toprule
    \textbf{Meta-prompt Attribute} & \textbf{LLM General Capabilities} & \textbf{Translation} \\
    \midrule			
        \multicolumn{3}{l}{\textbf{Topic}} \\
        Automotive industry & 1.20\% & 1.00\% \\
            \quad Vehicle manufacturing & 0.20\% & 0.00\% \\
            \quad Electric vehicles & 0.20\% & 0.40\% \\
            \quad Autonomous technology & 0.40\% & 0.40\% \\
            \quad Auto design & 0.00\% & 0.20\% \\
            \quad Car technology & 0.00\% & 0.00\% \\
            \quad Safety systems & 0.00\% & 0.00\% \\
            \quad Market trends & 0.40\% & 0.00\% \\
        Social media & 0.60\% & 1.80\% \\
            \quad Platform development & 0.00\% & 0.20\% \\
            \quad Content creation & 0.40\% & 0.40\% \\
            \quad Social analytics & 0.00\% & 0.20\% \\
            \quad User engagement & 0.00\% & 0.40\% \\
            \quad Social commerce & 0.00\% & 0.00\% \\
            \quad Digital communities & 0.00\% & 0.00\% \\
            \quad Influencer marketing & 0.40\% & 0.80\% \\
        Dating \& relationships & 0.40\% & 0.60\% \\
            \quad Online dating & 0.40\% & 0.00\% \\
            \quad Relationship counseling & 0.00\% & 0.20\% \\
            \quad Dating apps & 0.00\% & 0.20\% \\
            \quad Social connection & 0.00\% & 0.00\% \\
            \quad Relationship psychology & 0.00\% & 0.20\% \\
            \quad Dating culture & 0.00\% & 0.00\% \\
            \quad Personal growth & 0.00\% & 0.00\% \\
        Parenting \& family & 2.20\% & 0.80\% \\
            \quad Child development & 0.00\% & 0.00\% \\
            \quad Family health & 0.40\% & 0.00\% \\
            \quad Education & 1.00\% & 0.20\% \\
            \quad Family dynamics & 0.20\% & 0.20\% \\
            \quad Parenting resources & 0.20\% & 0.00\% \\
            \quad Child safety & 0.20\% & 0.40\% \\
            \quad Family planning & 0.40\% & 0.00\% \\
        Arts \& culture & 0.80\% & 2.20\% \\
            \quad Visual arts & 0.00\% & 0.00\% \\
            \quad Performance art & 0.00\% & 0.60\% \\
            \quad Cultural heritage & 1.60\% & 3.20\% \\
            \quad Art education & 0.00\% & 0.20\% \\
            \quad Digital art & 0.20\% & 0.20\% \\
            \quad Cultural events & 0.20\% & 0.20\% \\
            \quad Art market & 0.00\% & 0.20\% \\
        Music industry & 0.60\% & 0.40\% \\
            \quad Music production & 0.00\% & 0.00\% \\
            \quad Digital distribution & 0.20\% & 0.20\% \\
            \quad Live events & 0.40\% & 0.00\% \\
            \quad Music technology & 0.00\% & 0.00\% \\
            \quad Artist management & 0.00\% & 0.00\% \\
            \quad Music publishing & 0.00\% & 0.20\% \\
            \quad Industry trends & 0.00\% & 0.00\% \\
        Film \& cinema & 1.60\% & 1.00\% \\
            \quad Film production & 0.20\% & 0.20\% \\
            \quad Distribution & 0.00\% & 0.00\% \\
            \quad Film technology & 0.60\% & 0.20\% \\
            \quad Cinema innovation & 0.00\% & 0.00\% \\
            \quad Film industry & 0.40\% & 1.00\% \\
            \quad Digital effects & 0.80\% & 0.60\% \\
            \quad Film marketing & 0.20\% & 0.20\% \\
    \bottomrule
\end{tabular}
    \caption{
        Part 6 of all meta-prompt attribute values and their prevalence on all Zero-Shots Benchmarks released.
        Distributions are the same across languages and language pairs, respectively.
        The VLM task meta-prompt does not have attributes, only an image.
    }
    \label{tab:attrs-and-dists-6}
\end{table}

\begin{table}[H]
    \centering
    \setlength{\tabcolsep}{3pt}
\renewcommand{\arraystretch}{1.3}
\scriptsize
\begin{tabular}{l c c}
    \toprule
    \textbf{Meta-prompt Attribute} & \textbf{LLM General Capabilities} & \textbf{Translation} \\
    \midrule			
        \multicolumn{3}{l}{\textbf{Topic}} \\
        Books \& literature & 1.20\% & 1.00\% \\
            \quad Publishing industry & 0.40\% & 0.20\% \\
            \quad Digital publishing & 0.00\% & 0.20\% \\
            \quad Literary trends & 0.40\% & 0.20\% \\
            \quad Book marketing & 0.00\% & 0.20\% \\
            \quad Author platform & 0.00\% & 0.00\% \\
            \quad Literary events & 0.40\% & 0.20\% \\
            \quad Reading technology & 0.00\% & 0.00\% \\
        Food \& cuisine & 0.80\% & 1.80\% \\
            \quad Culinary arts & 1.20\% & 0.60\% \\
            \quad Food culture & 1.00\% & 1.20\% \\
            \quad Restaurant industry & 0.00\% & 0.60\% \\
            \quad Food innovation & 0.00\% & 0.20\% \\
            \quad Dietary trends & 0.00\% & 0.20\% \\
            \quad Food technology & 0.20\% & 0.40\% \\
            \quad Culinary education & 0.40\% & 0.20\% \\
        Sports \& recreation & 0.40\% & 0.80\% \\
            \quad Professional sports & 0.00\% & 0.20\% \\
            \quad Recreational activities & 0.00\% & 0.00\% \\
            \quad Sports technology & 0.20\% & 0.20\% \\
            \quad Fitness training & 0.00\% & 0.40\% \\
            \quad Sports medicine & 0.00\% & 0.60\% \\
            \quad Equipment innovation & 0.00\% & 0.00\% \\
            \quad Sports management & 0.20\% & 0.40\% \\
        Fitness \& wellness & 1.00\% & 2.00\% \\
            \quad Exercise programs & 0.20\% & 0.00\% \\
            \quad Wellness technology & 0.00\% & 0.60\% \\
            \quad Mental health & 0.20\% & 0.40\% \\
            \quad Nutrition & 0.00\% & 0.60\% \\
            \quad Personal training & 0.40\% & 0.00\% \\
            \quad Wellness education & 0.20\% & 0.20\% \\
            \quad Health tracking & 0.00\% & 0.20\% \\
        Mental health & 2.20\% & 1.80\% \\
            \quad Therapy services & 0.00\% & 0.80\% \\
            \quad Mental health technology & 0.00\% & 0.20\% \\
            \quad Support programs & 1.00\% & 0.40\% \\
            \quad Research \& treatment & 0.00\% & 0.20\% \\
            \quad Workplace mental health & 0.80\% & 0.20\% \\
            \quad Youth mental health & 0.20\% & 0.00\% \\
            \quad Mental health education & 0.20\% & 0.00\% \\
        Architecture \& design & 0.60\% & 1.80\% \\
            \quad Urban architecture & 0.20\% & 0.00\% \\
            \quad Sustainable design & 0.00\% & 0.40\% \\
            \quad Interior design & 0.00\% & 0.60\% \\
            \quad Digital architecture & 0.00\% & 0.00\% \\
            \quad Design innovation & 0.00\% & 0.20\% \\
            \quad Building technology & 0.20\% & 0.00\% \\
            \quad Architectural heritage & 0.20\% & 0.60\% \\
        Real estate & 2.20\% & 1.80\% \\
            \quad Property development & 0.40\% & 0.20\% \\
            \quad Real estate technology & 0.20\% & 0.20\% \\
            \quad Market analysis & 0.60\% & 0.20\% \\
            \quad Property management & 0.20\% & 0.00\% \\
            \quad Investment & 0.20\% & 0.20\% \\
            \quad Commercial real estate & 0.20\% & 0.40\% \\
            \quad Sustainable building & 0.40\% & 0.60\% \\
    \bottomrule
\end{tabular}
    \caption{
        Part 7 of all meta-prompt attribute values and their prevalence on all Zero-Shots Benchmarks released.
        Distributions are the same across languages and language pairs, respectively.
        The VLM task meta-prompt does not have attributes, only an image.
    }
    \label{tab:attrs-and-dists-7}
\end{table}

\begin{table}[H]
    \centering
    \setlength{\tabcolsep}{3pt}
\renewcommand{\arraystretch}{1.3}
\scriptsize
\begin{tabular}{l c c}
    \toprule
    \textbf{Meta-prompt Attribute} & \textbf{LLM General Capabilities} & \textbf{Translation} \\
    \midrule			
        \multicolumn{3}{l}{\textbf{Topic}} \\
        Telecommunications & 1.60\% & 1.20\% \\
            \quad Network infrastructure & 0.80\% & 0.20\% \\
            \quad Mobile technology & 0.20\% & 0.20\% \\
            \quad Communication services & 0.00\% & 0.60\% \\
            \quad Digital networks & 0.00\% & 0.00\% \\
            \quad Telecom innovation & 0.40\% & 0.20\% \\
            \quad Wireless technology & 0.00\% & 0.00\% \\
            \quad Industry standards & 0.20\% & 0.00\% \\
        Renewable energy & 1.80\% & 1.40\% \\
            \quad Solar power & 0.40\% & 0.00\% \\
            \quad Wind energy & 0.20\% & 0.20\% \\
            \quad Energy storage & 0.20\% & 0.40\% \\
            \quad Green technology & 0.20\% & 0.20\% \\
            \quad Energy policy & 0.40\% & 0.20\% \\
            \quad Sustainable development & 0.40\% & 0.20\% \\
            \quad Clean energy innovation & 0.00\% & 0.20\% \\
        Space exploration & 1.60\% & 1.20\% \\
            \quad Space technology & 0.00\% & 0.00\% \\
            \quad Satellite systems & 0.00\% & 0.20\% \\
            \quad Space research & 0.00\% & 0.20\% \\
            \quad Space industry & 0.20\% & 0.00\% \\
            \quad Astronomical discovery & 0.80\% & 0.40\% \\
            \quad Space travel & 0.20\% & 0.00\% \\
            \quad Space policy & 0.40\% & 0.40\% \\
        Wildlife \& nature & 1.20\% & 1.80\% \\
            \quad Conservation & 0.60\% & 0.40\% \\
            \quad Biodiversity & 0.00\% & 0.20\% \\
            \quad Wildlife research & 0.00\% & 0.00\% \\
            \quad Environmental protection & 0.20\% & 0.20\% \\
            \quad Natural habitats & 0.40\% & 0.20\% \\
            \quad Species preservation & 0.00\% & 0.40\% \\
            \quad Ecosystem management & 0.00\% & 0.40\% \\
        Weather \& climate & 1.20\% & 1.00\% \\
            \quad Climate science & 0.40\% & 0.80\% \\
            \quad Weather forecasting & 0.00\% & 0.40\% \\
            \quad Climate change & 0.20\% & 0.00\% \\
            \quad Atmospheric research & 0.20\% & 0.00\% \\
            \quad Environmental impact & 0.00\% & 0.00\% \\
            \quad Climate technology & 0.60\% & 0.20\% \\
            \quad Weather systems & 0.00\% & 0.20\% \\
        History \& heritage & 1.40\% & 1.20\% \\
            \quad Historical research & 0.20\% & 0.40\% \\
            \quad Cultural preservation & 0.20\% & 0.00\% \\
            \quad Archaeological studies & 0.40\% & 0.20\% \\
            \quad Heritage conservation & 0.20\% & 0.20\% \\
            \quad Historical education & 0.20\% & 0.00\% \\
            \quad Digital archives & 0.00\% & 0.20\% \\
            \quad Cultural memory & 0.20\% & 0.20\% \\
        Politics \& governance & 2.20\% & 1.20\% \\
            \quad Political systems & 0.20\% & 0.40\% \\
            \quad Public policy & 0.40\% & 0.40\% \\
            \quad Government innovation & 0.20\% & 0.00\% \\
            \quad Electoral processes & 0.80\% & 0.00\% \\
            \quad Political communication & 0.40\% & 0.20\% \\
            \quad Civic technology & 0.00\% & 0.20\% \\
            \quad Governance reform & 0.20\% & 0.00\% \\
    \bottomrule
\end{tabular}
    \caption{
        Part 8 of all meta-prompt attribute values and their prevalence on all Zero-Shots Benchmarks released.
        Distributions are the same across languages and language pairs, respectively.
        The VLM task meta-prompt does not have attributes, only an image.
    }
    \label{tab:attrs-and-dists-8}
\end{table}

\begin{table}[H]
    \centering
    \setlength{\tabcolsep}{3pt}
\renewcommand{\arraystretch}{1.3}
\scriptsize
\begin{tabular}{l c c}
    \toprule
    \textbf{Meta-prompt Attribute} & \textbf{LLM General Capabilities} & \textbf{Translation} \\
    \midrule			
        \multicolumn{3}{l}{\textbf{Topic}} \\
        Ngos \& nonprofits & 1.00\% & 1.40\% \\
            \quad Social impact & 0.00\% & 0.00\% \\
            \quad Humanitarian aid & 0.40\% & 0.40\% \\
            \quad Nonprofit management & 0.20\% & 0.00\% \\
            \quad Fundraising & 0.20\% & 0.00\% \\
            \quad Community development & 0.00\% & 0.20\% \\
            \quad International development & 0.00\% & 0.40\% \\
            \quad Social innovation & 0.20\% & 0.40\% \\
        New york city & 1.20\% & 1.40\% \\
            \quad Business \& finance & 0.00\% & 0.40\% \\
            \quad Arts \& culture & 0.20\% & 0.20\% \\
            \quad Food \& dining & 0.60\% & 0.40\% \\
            \quad Urban development & 1.40\% & 1.40\% \\
            \quad Tourism & 1.00\% & 0.60\% \\
            \quad Entertainment & 0.60\% & 0.80\% \\
            \quad Sports teams & 0.00\% & 0.20\% \\
        London & 0.60\% & 1.80\% \\
            \quad Financial services & 0.20\% & 0.00\% \\
            \quad Cultural heritage & 1.60\% & 3.20\% \\
            \quad Food scene & 0.80\% & 0.40\% \\
            \quad Theatre \& arts & 0.00\% & 0.40\% \\
            \quad Royal traditions & 0.40\% & 0.00\% \\
            \quad Urban transport & 0.00\% & 0.00\% \\
            \quad Sports culture & 0.60\% & 1.00\% \\
        Tokyo & 1.60\% & 0.80\% \\
            \quad Technology industry & 0.20\% & 0.60\% \\
            \quad Traditional culture & 0.40\% & 0.40\% \\
            \quad Cuisine & 0.00\% & 0.00\% \\
            \quad Pop culture & 1.40\% & 0.60\% \\
            \quad Urban innovation & 0.60\% & 0.80\% \\
            \quad Fashion & 0.40\% & 0.40\% \\
            \quad Entertainment districts & 0.40\% & 0.40\% \\
        Paris & 1.00\% & 1.40\% \\
            \quad Fashion industry & 0.20\% & 0.60\% \\
            \quad Culinary arts & 1.20\% & 0.60\% \\
            \quad Cultural landmarks & 0.20\% & 0.00\% \\
            \quad Art scene & 0.20\% & 0.20\% \\
            \quad Tourism industry & 0.80\% & 0.80\% \\
            \quad Luxury brands & 0.80\% & 0.80\% \\
            \quad Urban life & 0.60\% & 0.60\% \\
        Berlin & 2.20\% & 1.60\% \\
            \quad Startup scene & 0.40\% & 0.20\% \\
            \quad Cultural history & 0.00\% & 0.80\% \\
            \quad Nightlife & 0.40\% & 0.00\% \\
            \quad Art community & 0.40\% & 0.00\% \\
            \quad Tech industry & 1.20\% & 0.80\% \\
            \quad Urban planning & 1.00\% & 1.20\% \\
            \quad Alternative culture & 0.40\% & 0.20\% \\
        Singapore & 1.00\% & 1.00\% \\
            \quad Financial hub & 0.20\% & 0.40\% \\
            \quad Food culture & 1.00\% & 1.20\% \\
            \quad Smart city initiatives & 0.00\% & 0.00\% \\
            \quad Cultural diversity & 0.00\% & 0.20\% \\
            \quad Business innovation & 0.20\% & 1.20\% \\
            \quad Urban planning & 1.00\% & 1.20\% \\
            \quad Education & 1.00\% & 0.20\% \\
    \bottomrule
\end{tabular}
    \caption{
        Part 9 of all meta-prompt attribute values and their prevalence on all Zero-Shots Benchmarks released.
        Distributions are the same across languages and language pairs, respectively.
        The VLM task meta-prompt does not have attributes, only an image.
    }
    \label{tab:attrs-and-dists-9}
\end{table}

\begin{table}[H]
    \centering
    \setlength{\tabcolsep}{3pt}
\renewcommand{\arraystretch}{1.3}
\scriptsize
\begin{tabular}{l c c}
    \toprule
    \textbf{Meta-prompt Attribute} & \textbf{LLM General Capabilities} & \textbf{Translation} \\
    \midrule			
        \multicolumn{3}{l}{\textbf{Topic}} \\
        Dubai & 1.60\% & 1.00\% \\
            \quad Business center & 0.80\% & 0.20\% \\
            \quad Tourism industry & 0.80\% & 0.80\% \\
            \quad Luxury lifestyle & 0.20\% & 0.00\% \\
            \quad Urban development & 1.40\% & 1.40\% \\
            \quad International trade & 0.60\% & 0.20\% \\
            \quad Technology innovation & 0.20\% & 0.00\% \\
            \quad Cultural traditions & 0.60\% & 0.40\% \\
        São paulo & 0.80\% & 1.00\% \\
            \quad Business hub & 1.20\% & 1.40\% \\
            \quad Cultural scene & 0.20\% & 0.00\% \\
            \quad Food \& dining & 0.60\% & 0.40\% \\
            \quad Urban life & 0.60\% & 0.60\% \\
            \quad Entertainment & 0.60\% & 0.80\% \\
            \quad Fashion industry & 0.20\% & 0.60\% \\
            \quad Sports culture & 0.60\% & 1.00\% \\
        Sydney & 1.60\% & 0.60\% \\
            \quad Lifestyle \& culture & 0.20\% & 0.20\% \\
            \quad Business district & 0.00\% & 0.00\% \\
            \quad Tourism & 1.00\% & 0.60\% \\
            \quad Sports events & 0.20\% & 0.00\% \\
            \quad Entertainment & 0.60\% & 0.80\% \\
            \quad Food scene & 0.80\% & 0.40\% \\
            \quad Urban development & 1.40\% & 1.40\% \\
        Mumbai & 1.00\% & 2.40\% \\
            \quad Film industry & 0.40\% & 1.00\% \\
            \quad Business center & 0.80\% & 0.20\% \\
            \quad Cultural heritage & 1.60\% & 3.20\% \\
            \quad Food culture & 1.00\% & 1.20\% \\
            \quad Urban development & 1.40\% & 1.40\% \\
            \quad Entertainment & 0.60\% & 0.80\% \\
            \quad Fashion & 0.40\% & 0.40\% \\
        Madrid & 1.00\% & 1.20\% \\
            \quad Cultural heritage & 1.60\% & 3.20\% \\
            \quad Food \& wine & 1.20\% & 0.60\% \\
            \quad Sports culture & 0.60\% & 1.00\% \\
            \quad Arts scene & 1.00\% & 0.40\% \\
            \quad Business hub & 1.20\% & 1.40\% \\
            \quad Urban life & 0.60\% & 0.60\% \\
            \quad Tourism & 1.00\% & 0.60\% \\
        Lisbon & 0.80\% & 1.80\% \\
            \quad Startup ecosystem & 0.20\% & 0.00\% \\
            \quad Cultural history & 0.00\% & 0.80\% \\
            \quad Food \& wine & 1.20\% & 0.60\% \\
            \quad Tourism industry & 0.80\% & 0.80\% \\
            \quad Urban innovation & 0.60\% & 0.80\% \\
            \quad Maritime heritage & 0.00\% & 0.20\% \\
            \quad Arts scene & 1.00\% & 0.40\% \\
        Stockholm & 0.80\% & 1.40\% \\
            \quad Tech innovation & 0.20\% & 0.00\% \\
            \quad Design culture & 0.20\% & 0.00\% \\
            \quad Sustainability & 0.20\% & 0.00\% \\
            \quad Urban planning & 1.00\% & 1.20\% \\
            \quad Business hub & 1.20\% & 1.40\% \\
            \quad Cultural scene & 0.20\% & 0.00\% \\
            \quad Food \& lifestyle & 0.00\% & 0.20\% \\
    \bottomrule
\end{tabular}
    \caption{
        Part 10 of all meta-prompt attribute values and their prevalence on all Zero-Shots Benchmarks released.
        Distributions are the same across languages and language pairs, respectively.
        The VLM task meta-prompt does not have attributes, only an image.
    }
    \label{tab:attrs-and-dists-10}
\end{table}

\begin{table}[H]
    \centering
    \setlength{\tabcolsep}{3pt}
\renewcommand{\arraystretch}{1.3}
\scriptsize
\begin{tabular}{l c c}
    \toprule
    \textbf{Meta-prompt Attribute} & \textbf{LLM General Capabilities} & \textbf{Translation} \\
    \midrule			
        \multicolumn{3}{l}{\textbf{Topic}} \\
        Amsterdam & 1.20\% & 0.80\% \\
            \quad Cultural heritage & 1.60\% & 3.20\% \\
            \quad Urban planning & 1.00\% & 1.20\% \\
            \quad Business innovation & 0.20\% & 1.20\% \\
            \quad Art scene & 0.20\% & 0.20\% \\
            \quad Cycling culture & 0.40\% & 0.00\% \\
            \quad Tourism & 1.00\% & 0.60\% \\
            \quad Tech industry & 1.20\% & 0.80\% \\
        Seoul & 2.40\% & 1.80\% \\
            \quad Tech industry & 1.20\% & 0.80\% \\
            \quad Pop culture & 1.40\% & 0.60\% \\
            \quad Food scene & 0.80\% & 0.40\% \\
            \quad Fashion trends & 0.60\% & 0.40\% \\
            \quad Urban innovation & 0.60\% & 0.80\% \\
            \quad Entertainment & 0.60\% & 0.80\% \\
            \quad Business hub & 1.20\% & 1.40\% \\
        Japan & 1.00\% & 1.40\% \\
            \quad Traditional culture & 0.40\% & 0.40\% \\
            \quad Technology industry & 0.20\% & 0.60\% \\
            \quad Popular culture & 0.00\% & 0.20\% \\
            \quad Business practices & 0.20\% & 0.20\% \\
            \quad Food \& cuisine & 0.40\% & 0.20\% \\
            \quad Arts \& crafts & 0.20\% & 0.00\% \\
            \quad Social customs & 0.00\% & 0.40\% \\
        France & 1.80\% & 1.00\% \\
            \quad Culinary arts & 1.20\% & 0.60\% \\
            \quad Fashion industry & 0.20\% & 0.60\% \\
            \quad Cultural heritage & 1.60\% & 3.20\% \\
            \quad Wine industry & 0.40\% & 0.20\% \\
            \quad Tourism & 1.00\% & 0.60\% \\
            \quad Business culture & 0.60\% & 0.40\% \\
            \quad Arts \& literature & 0.40\% & 0.00\% \\
        Germany & 1.00\% & 1.80\% \\
            \quad Automotive industry & 0.40\% & 0.20\% \\
            \quad Business innovation & 0.20\% & 1.20\% \\
            \quad Cultural traditions & 0.60\% & 0.40\% \\
            \quad Sports culture & 0.60\% & 1.00\% \\
            \quad Technology sector & 0.20\% & 0.00\% \\
            \quad Education system & 0.20\% & 0.20\% \\
            \quad Urban development & 1.40\% & 1.40\% \\
        Brazil & 1.20\% & 0.80\% \\
            \quad Cultural festivals & 0.40\% & 1.00\% \\
            \quad Sports culture & 0.60\% & 1.00\% \\
            \quad Business environment & 0.20\% & 0.60\% \\
            \quad Food \& cuisine & 0.40\% & 0.20\% \\
            \quad Music scene & 0.00\% & 0.40\% \\
            \quad Tourism industry & 0.80\% & 0.80\% \\
            \quad Urban life & 0.60\% & 0.60\% \\
        India & 1.40\% & 1.20\% \\
            \quad Technology sector & 0.20\% & 0.00\% \\
            \quad Cultural diversity & 0.00\% & 0.20\% \\
            \quad Film industry & 0.40\% & 1.00\% \\
            \quad Culinary traditions & 0.40\% & 0.00\% \\
            \quad Business hub & 1.20\% & 1.40\% \\
            \quad Traditional arts & 0.00\% & 0.20\% \\
            \quad Festival culture & 0.00\% & 0.20\% \\
    \bottomrule
\end{tabular}
    \caption{
        Part 11 of all meta-prompt attribute values and their prevalence on all Zero-Shots Benchmarks released.
        Distributions are the same across languages and language pairs, respectively.
        The VLM task meta-prompt does not have attributes, only an image.
    }
    \label{tab:attrs-and-dists-11}
\end{table}

\begin{table}[H]
    \centering
    \setlength{\tabcolsep}{3pt}
\renewcommand{\arraystretch}{1.3}
\scriptsize
\begin{tabular}{l c c}
    \toprule
    \textbf{Meta-prompt Attribute} & \textbf{LLM General Capabilities} & \textbf{Translation} \\
    \midrule			
        \multicolumn{3}{l}{\textbf{Topic}} \\
        Italy & 1.40\% & 0.80\% \\
            \quad Food \& wine & 1.20\% & 0.60\% \\
            \quad Fashion industry & 0.20\% & 0.60\% \\
            \quad Cultural heritage & 1.60\% & 3.20\% \\
            \quad Tourism sector & 0.20\% & 0.40\% \\
            \quad Design industry & 0.00\% & 0.20\% \\
            \quad Business culture & 0.60\% & 0.40\% \\
            \quad Arts scene & 1.00\% & 0.40\% \\
        Spain & 1.40\% & 0.40\% \\
            \quad Cultural traditions & 0.60\% & 0.40\% \\
            \quad Culinary arts & 1.20\% & 0.60\% \\
            \quad Tourism industry & 0.80\% & 0.80\% \\
            \quad Sports culture & 0.60\% & 1.00\% \\
            \quad Business environment & 0.20\% & 0.60\% \\
            \quad Arts \& entertainment & 0.40\% & 0.20\% \\
            \quad Urban life & 0.60\% & 0.60\% \\
        China & 1.60\% & 1.20\% \\
            \quad Technology industry & 0.20\% & 0.60\% \\
            \quad Business culture & 0.60\% & 0.40\% \\
            \quad Traditional customs & 0.40\% & 0.20\% \\
            \quad Urban development & 1.40\% & 1.40\% \\
            \quad Cultural heritage & 1.60\% & 3.20\% \\
            \quad Food culture & 1.00\% & 1.20\% \\
            \quad Innovation hub & 0.00\% & 0.00\% \\
        United kingdom & 1.00\% & 1.20\% \\
            \quad Financial services & 0.20\% & 0.00\% \\
            \quad Cultural heritage & 1.60\% & 3.20\% \\
            \quad Education system & 0.20\% & 0.20\% \\
            \quad Business innovation & 0.20\% & 1.20\% \\
            \quad Arts \& entertainment & 0.40\% & 0.20\% \\
            \quad Sports culture & 0.60\% & 1.00\% \\
            \quad Urban life & 0.60\% & 0.60\% \\
        Portugal & 1.00\% & 0.60\% \\
            \quad Tourism industry & 0.80\% & 0.80\% \\
            \quad Food \& wine & 1.20\% & 0.60\% \\
            \quad Cultural heritage & 1.60\% & 3.20\% \\
            \quad Business innovation & 0.20\% & 1.20\% \\
            \quad Maritime culture & 0.20\% & 0.20\% \\
            \quad Urban development & 1.40\% & 1.40\% \\
            \quad Arts scene & 1.00\% & 0.40\% \\
        Poetry & 0.60\% & 0.60\% \\
            \quad Modernism & 0.20\% & 0.00\% \\
            \quad Contemporary & 0.40\% & 0.00\% \\
            \quad Modernism & 0.20\% & 0.00\% \\
            \quad Haiku & 0.00\% & 0.40\% \\
            \quad European poetry & 0.00\% & 0.00\% \\
            \quad Asian poetry & 0.00\% & 0.00\% \\
            \quad Theme identification & 0.00\% & 0.20\% \\
    \bottomrule
\end{tabular}
    \caption{
        Part 12 of all meta-prompt attribute values and their prevalence on all Zero-Shots Benchmarks released.
        Distributions are the same across languages and language pairs, respectively.
        The VLM task meta-prompt does not have attributes, only an image.
    }
    \label{tab:attrs-and-dists-12}
\end{table}

\begin{table}[H]
    \centering
    \setlength{\tabcolsep}{3pt}
\renewcommand{\arraystretch}{1.3}
\scriptsize
\begin{tabular}{l c c}
    \toprule
    \textbf{Meta-prompt Attribute} & \textbf{LLM General Capabilities} & \textbf{Translation} \\
    \midrule			
        \multicolumn{3}{l}{\textbf{Difficulty}} \\
        Easy & 21.20\% & - \\
        Medium & 20.40\% & - \\
        Hard & 18.80\% & - \\
        Very hard & 18.20\% & - \\
        Arguably impossible to answer & 21.40\% & - \\
    \cdashlinelr{1-3} \\
    \multicolumn{3}{l}{\textbf{Style}} \\
        Creative & 7.20\% & 5.60\% \\
        Concise & 5.40\% & 5.80\% \\
        Technical & 6.00\% & 5.20\% \\
        Formal & 6.40\% & 5.80\% \\
        Informal & 4.60\% & 6.40\% \\
        Narrative & 5.60\% & 5.00\% \\
        Persuasive & 4.40\% & 5.60\% \\
        Descriptive & 6.20\% & 6.20\% \\
        Analytical & 5.40\% & 7.20\% \\
        Humorous & 7.40\% & 5.20\% \\
        Poetic & 5.40\% & 6.40\% \\
        Casual & 5.60\% & 5.60\% \\
        Academic & 5.40\% & 3.40\% \\
        Journalistic & 4.00\% & 5.20\% \\
        Neutral & 3.60\% & 5.40\% \\
        Elaborate & 5.60\% & 5.60\% \\
        Minimalist & 6.80\% & 5.60\% \\
        Rushed & 5.00\% & 4.80\% \\
    \cdashlinelr{1-3} \\
    \multicolumn{3}{l}{\textbf{Writer}} \\
        Children & 9.40\% & - \\
        Teenagers & 9.60\% & - \\
        Young adults & 8.80\% & - \\
        Parents & 7.20\% & - \\
        Seniors & 6.00\% & - \\
        Professionals & 6.00\% & - \\
        College students & 9.20\% & - \\
        Educators & 10.20\% & - \\
        General public & 10.00\% & - \\
        Beginners & 8.00\% & - \\
        Experts & 6.80\% & - \\
        Middle-aged adults & 8.80\% & - \\
    \cdashlinelr{1-3} \\
    \multicolumn{3}{l}{\textbf{Writing Proficiency}} \\
        Low & 24.20\% & - \\
        Medium & 23.00\% & - \\
        High & 25.60\% & - \\
        Virtuoso & 27.20\% & - \\
    \cdashlinelr{1-3} \\
    \multicolumn{3}{l}{\textbf{Length}} \\
        Short & 25.60\% & 25.60\% \\
        Medium & 25.80\% & 23.40\% \\
        Long & 23.40\% & 26.20\% \\
        Very long & 25.20\% & 24.80\% \\

    \bottomrule
\end{tabular}
    \caption{
        Part 13 of all meta-prompt attribute values and their prevalence on all Zero-Shots Benchmarks released.
        Distributions are the same across languages and language pairs, respectively.
        The VLM task meta-prompt does not have attributes, only an image.
    }
    \label{tab:attrs-and-dists-13}
\end{table}

\subsection{ZSB Judgement Prompts}\label{subsec:apx-judge-prompts}

\begin{figure}[H]
    \small
    \renewcommand\tabularxcolumn[1]{m{#1}}
   \centering
\begin{tabularx}{\linewidth}{*{1}{>{\arraybackslash}X}}
You are an expert judge evaluating response quality for prompts on a variety of topics. You will be presented with: \\
\hfill \\
- An original prompt \\
- A response to evaluate \\
\hfill \\
Rate the response on a scale of 1-6 based on these key criteria: \\
\hfill \\
- Accuracy and factual correctness \\
- Comprehensiveness of the answer \\
- Clarity and organization \\
- Relevance to the prompt \\
- Supporting evidence/examples (if applicable) \\
- Objectivity and balance (if applicable) \\
\hfill \\
Scoring Rubric: \\
6 - Outstanding \\
\hfill \\
- Exceptionally accurate and well-researched information \\
- Comprehensive coverage with valuable additional context \\
- Perfectly structured and crystal clear explanation \\
- Directly addresses all aspects of the prompt \\
- Strong supporting evidence and examples \\
- Maintains perfect objectivity where needed \\
\hfill \\
5 - Excellent \\
\hfill \\
- Highly accurate information \\
- Thorough coverage of the topic \\
- Very clear and well-organized \\
- Closely aligned with the prompt \\
- Good supporting evidence \\
- Strong objectivity \\
\hfill \\
4 - Good \\
\hfill \\
- Generally accurate information \\
- Adequate coverage of main points \\
- Clear organization with minor issues \\
- Mostly relevant to the prompt \\
- Basic supporting evidence \\
- Reasonable objectivity \\
\hfill \\
3 - Fair \\
\hfill \\
- Some inaccuracies present \\
- Incomplete coverage \\
- Organization needs improvement \\
- Partially misses prompt intent \\
- Limited supporting evidence \\
- Some bias present \\
\hfill \\
\end{tabularx}
\caption{ZSB judge prompt for general capabilities part 1.}\label{fig:general_capabilities_judge_prompt_1}
\end{figure}

\begin{figure}[H]
    \small
    \renewcommand\tabularxcolumn[1]{m{#1}}
   \centering
\begin{tabularx}{\linewidth}{*{1}{>{\arraybackslash}X}}
2 - Poor \\
\hfill \\
- Multiple inaccuracies \\
- Superficial coverage \\
- Unclear organization \\
- Largely misses prompt intent \\
- Minimal supporting evidence \\
- Clear bias present \\
\hfill \\
1 - Inadequate \\
\hfill \\
- Fundamentally incorrect information \\
- Severely incomplete \\
- Disorganized or incomprehensible \\
- Fails to address prompt \\
- No supporting evidence \\
- Extremely biased or misleading \\
\hfill \\
Provide your evaluation in this JSON format: \\
\hfill \\
\{``feedback'': ``$<$detailed explanation of the score based on the criteria$>$'', ``result'': ``$<$only a number from 1 to 6$>$''\} \\
\hfill \\
Remember that a good response should be accurate, comprehensive, clear, and directly address the prompt while providing appropriate supporting evidence and maintaining objectivity where needed. Remember that a long response is not necessarily good. If the response is not in the language of the prompt, give it a score of 1. \\
\hfill \\
$<$START OF PROMPT$>$ \\
\$\{prompt\} \\
$<$END OF PROMPT$>$ \\
\hfill \\
$<$START OF ANSWER$>$ \\
\$\{answer\} \\
$<$END OF ANSWER$>$ \\
\hfill \\
You may proceed to evaluate the response. Ensure the output is valid JSON, without additional formatting or explanations.
\end{tabularx}
\caption{ZSB judge prompt for general capabilities part 2.}\label{fig:general_capabilities_judge_prompt_2}
\end{figure}


\begin{figure}[H]
    \small
    \renewcommand\tabularxcolumn[1]{m{#1}}
   \centering
\begin{tabularx}{\linewidth}{*{1}{>{\arraybackslash}X}}
You are an expert judge evaluating translation quality. You will be presented with: \\
\hfill \\
- An original text \\
- A translation to evaluate \\
\hfill \\
Rate the translation on a scale of 1-6 based on these key criteria: \\
\hfill \\
- Accuracy and fidelity to source \\
- Grammar and language correctness \\
- Natural flow and readability \\
- Terminology consistency \\
- Completeness of translation \\
- Technical precision \\
\hfill \\
Scoring Rubric: \\
6 - Outstanding \\
\hfill \\
- Perfect accuracy with source meaning \\
- Flawless grammar and language use \\
- Reads completely naturally in target language \\
- Consistent and precise terminology \\
- Complete translation with no omissions \\
- Excellent technical accuracy \\
\hfill \\
5 - Excellent \\
\hfill \\
- Very accurate rendering of source \\
- Strong grammar with minimal issues \\
- Natural-sounding translation \\
- Good terminology consistency \\
- Nearly complete coverage \\
- Strong technical accuracy \\
\hfill \\
4 - Good \\
\hfill \\
- Generally accurate translation \\
- Mostly correct grammar \\
- Readable with some awkward passages \\
- Generally consistent terminology \\
- Minor omissions only \\
- Adequate technical accuracy \\
\hfill \\
3 - Fair \\
\hfill \\
- Some accuracy issues \\
- Notable grammar problems \\
- Often unnatural phrasing \\
- Inconsistent terminology \\
- Several omissions \\
- Technical inaccuracies present \\
\end{tabularx}
\caption{ZSB judge prompt for translation part 1.}\label{fig:translation_judge_prompt_1}
\end{figure}

\begin{figure}[H]
    \small
    \renewcommand\tabularxcolumn[1]{m{#1}}
   \centering
\begin{tabularx}{\linewidth}{*{1}{>{\arraybackslash}X}}
2 - Poor \\
\hfill \\
- Significant accuracy issues \\
- Frequent grammar errors \\
- Unnatural throughout \\
- Poor terminology consistency \\
- Major omissions \\
- Many technical errors \\
\hfill \\
1 - Inadequate \\
\hfill \\
- Fails to convey source meaning \\
- Severe grammar issues \\
- Incomprehensible in target language \\
- No terminology consistency \\
- Incomplete translation \\
- Technical meaning lost \\
\hfill \\
Provide your evaluation in this JSON format:\\
\hfill \\
\{``feedback'': ``$<$detailed explanation of the score based on the criteria$>$'', ``result'': ``$<$only a number from 1 to 6$>$''\} \\
\hfill \\
$<$START OF SOURCE TEXT$>$ \\
\$\{prompt\} \\
$<$END OF SOURCE TEXT$>$ \\
\hfill \\
$<$START OF TRANSLATION$>$ \\
\$\{answer\} \\
$<$END OF TRANSLATION$>$ \\
\hfill \\                               
You may proceed to evaluate the translation. Ensure the output is valid JSON, without additional formatting or explanations.
\end{tabularx}
\caption{ZSB judge prompt for translation part 2.}\label{fig:translation_judge_prompt_2}
\end{figure}


\begin{figure}[H]
    \small
    \renewcommand\tabularxcolumn[1]{m{#1}}
   \centering
\begin{tabularx}{\linewidth}{*{1}{>{\arraybackslash}X}}
You are an expert judge evaluating how well vision language models analyze and respond to image-based prompts. You will be presented with: \\
\hfill \\
- The original prompt given to the model \\
- The model's response \\
\hfill \\
Rate the response on a scale of 1-6 based on these key criteria: \\
\hfill \\
- Visual Detail Recognition: How well does the response identify and describe specific elements in the image? \
- Reasoning \& Analysis: Does the response make meaningful connections and draw logical conclusions? \
- Prompt Adherence: How well does the response address all aspects of the prompt? \
- Accuracy: Is the information provided accurate based on what's visible in the image? \
- Depth of Understanding: Does the response demonstrate deeper insight beyond surface-level observation? \
\hfill \\
Scoring Rubric: \\
\hfill \\
6 - Outstanding \\
\hfill \\
- Exceptional attention to visual details \\
- Sophisticated reasoning and connections \\
- Addresses all prompt aspects comprehensively \\
- Perfect accuracy in image interpretation \\
- Deep insights beyond obvious observations \\
\hfill \\
5 - Excellent \\
\hfill \\
- Strong attention to visual details \\
- Clear reasoning and meaningful connections \\
- Addresses most prompt aspects well \\
- High accuracy in image interpretation \\
- Good contextual understanding \\
\hfill \\
4 - Good \\
\hfill \\
- Adequate visual detail recognition \\
- Basic but logical reasoning \\
- Addresses main prompt points \\
- Generally accurate interpretation \\
- Some contextual understanding \\
\hfill \\
3 - Fair \\
\hfill \\
- Missing some key visual details \\
- Limited reasoning or connections \\
- Partially addresses the prompt \\
- Some inaccuracies in interpretation \\
- Superficial understanding \\
\end{tabularx}
\caption{ZSB judge prompt for VLM general capabilities part 1.}\label{fig:vlm_general_capabilities_judge_prompt_1}
\end{figure}

\begin{figure}[H]
    \small
    \renewcommand\tabularxcolumn[1]{m{#1}}
   \centering
\begin{tabularx}{\linewidth}{*{1}{>{\arraybackslash}X}}
2 - Poor \\
\hfill \\
- Misses many visual details \\
- Weak or illogical reasoning \\
- Largely misses prompt intent \\
- Multiple inaccuracies \\
- Very surface-level analysis \\
\hfill \\
1 - Inadequate \\
\hfill \\
- Fails to identify basic visual elements \\
- No meaningful reasoning \\
- Doesn't address prompt \\
- Fundamentally incorrect \\
- No real understanding demonstrated \\
\hfill \\
Provide your evaluation in this JSON format: \\
\{``feedback'': ``''$<$detailed explanation of the score based on the criteria$>$'', ``result'': ``$<$only a number from 1 to 6$>$''\}
\hfill \\
Base your evaluation primarily on how well the response demonstrates visual understanding and reasoning capabilities compared to the reference answer. Consider both what is said and what should have been noticed but wasn't. \\
\hfill \\
$<$START OF PROMPT$>$
\$\{prompt\}
$<$END OF PROMPT$>$
\hfill \\
$<$START OF ANSWER$>$ \\
\$\{answer\} \\
$<$END OF ANSWER$>$ \\
\hfill \\
You may proceed to evaluate the response. Ensure the output is valid JSON, without additional formatting or explanations.
\end{tabularx}
\caption{ZSB judge prompt for VLM general capabilities part 2.}\label{fig:vlm_general_capabilities_judge_prompt_2}
\end{figure}

\subsection{ZSB Generated Data Examples}\label{subsec:apx-examples}
\begin{figure}[H]
    \small
    \renewcommand\tabularxcolumn[1]{m{#1}}
   \centering
\begin{tabularx}{\linewidth}{*{1}{>{\arraybackslash}X}}
Explain how the rise of electric vehicles is disrupting traditional automotive supply chains and manufacturing processes. Include specific examples of how this transition affects tier-1 suppliers, and predict which current automotive technologies might become obsolete by 2030. Support your analysis with recent market trends.
\end{tabularx}
\caption{ZSB Example for English general capabilities generated by \claudethreefive{}.}\label{fig:general_capabilities_example}
\end{figure}

\begin{figure}[H]
    \small
    \renewcommand\tabularxcolumn[1]{m{#1}}
   \centering
\begin{tabularx}{\linewidth}{*{1}{>{\arraybackslash}X}}
Dans le contexte de la transformation numérique de l'éducation en France, proposez un modèle théorique qui résoudrait simultanément les trois paradoxes suivants : (1) comment garantir une égalité d'accès aux ressources numériques tout en permettant une personnalisation totale de l'apprentissage, (2) comment maintenir l'interaction sociale essentielle au développement tout en maximisant les avantages de l'apprentissage autonome en ligne, et (3) comment assurer une évaluation objective des compétences acquises numériquement tout en respectant les différents rythmes d'apprentissage individuels. Votre réponse doit être applicable dans le système éducatif français actuel, avec les contraintes budgétaires existantes.
\end{tabularx}
\caption{ZSB Example for French general capabilities generated by \claudethreefive{}.}\label{fig:general_capabilities_example_french}
\end{figure}

\begin{figure}[H]
    \small
    \renewcommand\tabularxcolumn[1]{m{#1}}
   \centering
\begin{tabularx}{\linewidth}{*{1}{>{\arraybackslash}X}}
Translate to Czech. \\
\hfill \\
Primary source analysis remains fundamental to historical research methodology. Historians must carefully examine contemporary documents, artifacts, and records to construct accurate interpretations of past events. This process requires not only analytical skills but also an understanding of the historical context in which these sources were created. Critical evaluation of source authenticity and reliability continues to be essential in modern historical scholarship.
\end{tabularx}
\caption{ZSB Example for English-Czech translation generated by \claudethreefive{}.}\label{fig:translation_example}
\end{figure}

\begin{figure}[H]
    \small
    \renewcommand\tabularxcolumn[1]{m{#1}}
   \centering
\begin{tabularx}{\linewidth}{*{1}{>{\arraybackslash}X}}
Analyze this craft beer label's artistic and marketing elements. How do the visual design, name choice, and imagery work together to create a specific brand identity? What demographic might this beer be targeting, and what does the artistic style suggest about the brewery's positioning in the craft beer market?
\end{tabularx}
\begin{figure}
\centering
\includegraphics[width=0.25\linewidth]{zsb_examples/zsb_vlm_img.jpg}
\end{figure}
\caption{ZSB Example for VLM English general capabilities generated by \claudethreefive{}. The image is passed to the model somewhere else in its chat template.}\label{fig:vlm_general_capabilities_example}
\end{figure}

\subsection{ZSB Safety Metadata}\label{subsec:safety-metadata}
\begin{figure}[H]
    \small
    \renewcommand\tabularxcolumn[1]{m{#1}}
   \centering
\begin{tabularx}{\linewidth}{*{1}{>{\arraybackslash}X}}
    You are an expert safety auditor evaluating prompt safety for a variety of tasks. You will be presented with: \\
    \hfill \\
    - A prompt to evaluate \\
    \hfill \\
    Rate the prompt on a scale of 1-6 based on these key safety criteria: \\
    \hfill \\
    - Potential for harmful outputs (misinformation, illegal content, etc.) \\
    - Risk of personal data extraction \\
    - Potential for manipulation/social engineering \\
    - Possibility of circumventing safety guardrails \\
    - Appropriateness of content (violence, explicit material, etc.) \\
    - Overall ethical considerations \\
    \hfill \\
    Scoring Rubric: \\
    6 - Highly Safe \\
    \hfill \\
    - Appears to have minimal risk of generating harmful outputs \\
    - Unlikely to extract personal information \\
    - Very low manipulation potential \\
    - Does not attempt to circumvent safety mechanisms \\
    - Content seems appropriate for general audiences \\
    - Aligns well with ethical standards \\
    - Note: All prompts carry some inherent risk \\
    \hfill \\
    5 - Generally Safe \\
    \hfill \\
    - Low risk of generating harmful content \\
    - Minimal personal data collection concerns \\
    - Low manipulation potential \\
    - Respects standard safety boundaries \\
    - Content generally appropriate \\
    - Few ethical concerns identified \\
    \hfill \\
    4 - Moderately Safe \\
    \hfill \\
    - Some potential for problematic outputs \\
    - Minor personal data concerns \\
    - Some manipulation possibilities \\
    - Minor attempts to test boundaries \\
    - Some content concerns for certain audiences \\
    - Some ethical questions arise \\
    \hfill \\
    3 - Somewhat Risky \\
    \hfill \\
    - Moderate risk of generating harmful content \\
    - Clear attempts to collect personal information \\
    - Noticeable manipulation tactics \\
    - Attempts to work around safety measures \\
    - Content inappropriate for some audiences \\
    - Several ethical concerns present \\
\end{tabularx}
\caption{Prompt used to obtain safety metadata for all text-only ZSB datasets (part 1).}\label{fig:llm_safety_prompt_1}
\end{figure}

\begin{figure}[H]
    \small
    \renewcommand\tabularxcolumn[1]{m{#1}}
   \centering
\begin{tabularx}{\linewidth}{*{1}{>{\arraybackslash}X}}
    2 - Risky \\
    \hfill \\
    - High likelihood of harmful outputs \\
    - Explicit attempts to extract personal data \\
    - Strong manipulation techniques \\
    - Clear attempts to circumvent safety measures \\
    - Generally inappropriate content \\
    - Serious ethical problems identified \\
    \hfill \\
    1 - Highly Problematic \\
    \hfill \\
    - Explicitly designed to generate harmful content \\
    - Aggressive personal data extraction \\
    - Severe manipulation/social engineering \\
    - Direct attacks on safety mechanisms \\
    - Clearly inappropriate/illegal content requests \\
    - Fundamentally unethical in nature \\
    \hfill \\
    $<$START OF PROMPT$>$ \\
    \$\{prompt\} \\
    $<$END OF PROMPT$>$ \\
    \hfill \\
    Provide your evaluation in this JSON format: \\
    \{``feedback'': ``$<$detailed explanation of the score based on the criteria$>$'', ``result'': ``$<$only a number from 1 to 6$>$''\} \\
    Remember that safety evaluation requires careful consideration of potential misuse, unintended consequences, and the context in which the prompt might be used. Consider both obvious and subtle risks when providing your assessment. \\
    \hfill \\
    You may proceed to evaluate the response. Ensure the output is valid JSON, without additional formatting or explanations.
\end{tabularx}
\caption{Prompt used to obtain safety metadata for all text-only ZSB datasets (part 2).}\label{fig:llm_safety_prompt_2}
\end{figure}


\begin{figure}[H]
    \small
    \renewcommand\tabularxcolumn[1]{m{#1}}
   \centering
\begin{tabularx}{\linewidth}{*{1}{>{\arraybackslash}X}}
    You are an expert safety auditor evaluating prompt safety for multimodal tasks. You will be presented with: \\
    \hfill \\
    - A text prompt to evaluate \\
    - An accompanying image \\
    \hfill \\
    Rate the prompt-image combination on a scale of 1-6 based on these key safety criteria: \\
    \hfill \\
    - Potential for harmful outputs (misinformation, illegal content, etc.) \\
    - Risk of personal data extraction from text or image \\
    - Potential for manipulation/social engineering \\
    - Possibility of circumventing safety guardrails \\
    - Appropriateness of content (violence, explicit material, etc.) \\
    - Contextual risks when text and image interact \\
    - Overall ethical considerations \\
    \hfill \\
    Scoring Rubric: \\
    6 - Highly Safe \\
    \hfill \\
    - Text and image appear to have minimal risk of generating harmful outputs \\
    - Unlikely to extract personal information from either component \\
    - Very low manipulation potential in the multimodal context \\
    - Does not attempt to circumvent safety mechanisms \\
    - Both text and image seem appropriate for general audiences \\
    - Image-text combination does not create new risks \\
    - Aligns well with ethical standards \\
    - Note: All prompts carry some inherent risk \\
    \hfill \\
    5 - Generally Safe \\
    \hfill \\
    - Low risk of generating harmful content in either medium \\
    - Minimal personal data collection concerns \\
    - Low manipulation potential in multimodal context \\
    - Respects standard safety boundaries \\
    - Content generally appropriate in both text and image \\
    - Limited potential for harmful interactions between text and image \\
    - Few ethical concerns identified \\
    \hfill \\
    4 - Moderately Safe \\
    \hfill \\
    - Some potential for problematic outputs \\
    - Minor personal data concerns in text or visual elements \\
    - Some manipulation possibilities enhanced by multimedia combination \\
    - Minor attempts to test boundaries \\
    - Some content concerns for certain audiences in either component \\
    - Text-image pairing may create moderate contextual concerns \\
    - Some ethical questions arise \\
    \hfill \\
    3 - Somewhat Risky \\
    \hfill \\
    - Moderate risk of generating harmful content \\
    - Clear attempts to collect personal information via text or image \\
    - Noticeable manipulation tactics potentially strengthened by visual context \\
    - Attempts to work around safety measures \\
    - Either text or image inappropriate for some audiences \\
    - Multimodal combination creates new concerning interpretations \\
    - Several ethical concerns present \\
\end{tabularx}
\caption{Prompt used to obtain safety metadata for all VLM ZSB datasets (part 1).}\label{fig:vlm_safety_prompt_1}
\end{figure}

\begin{figure}[H]
    \small
    \renewcommand\tabularxcolumn[1]{m{#1}}
   \centering
\begin{tabularx}{\linewidth}{*{1}{>{\arraybackslash}X}}
    2 - Risky \\
    \hfill \\
    - High likelihood of harmful outputs \\
    - Explicit attempts to extract personal data from text or visual cues \\
    - Strong manipulation techniques utilizing both modalities \\
    - Clear attempts to circumvent safety measures \\
    - Generally inappropriate content in text, image, or their combination \\
    - Image significantly amplifies risks in text prompt or vice versa \\
    - Serious ethical problems identified \\
    \hfill \\
    1 - Highly Problematic \\
    \hfill \\
    - Explicitly designed to generate harmful content across modalities \\
    - Aggressive personal data extraction using text and visual elements \\
    - Severe manipulation/social engineering enhanced by multimedia approach \\
    - Direct attacks on safety mechanisms \\
    - Clearly inappropriate/illegal content in text, image, or their interaction \\
    - Text and image combination designed to bypass safety filters \\
    - Fundamentally unethical in nature \\
    \hfill \\
    $<$START OF PROMPT$>$ \\
    \$\{prompt\} \\
    $<$END OF PROMPT$>$ \\
    \hfill \\
    Provide your evaluation in this JSON format: \\
    \{``feedback'': ``$<$detailed explanation of the score based on the criteria$>$'', ``result'': ``$<$only a number from 1 to 6$>$''\} \\
    \hfill \\
    Remember that multimodal safety evaluation requires considering how text and images interact to potentially create new risks beyond what either component presents alone. Consider both obvious and subtle risks when providing your assessment, including how visual context might alter the interpretation of text and vice versa. \\
    \hfill \\
    You may proceed to evaluate the response. Ensure the output is valid JSON, without additional formatting or explanations.
\end{tabularx}
\caption{Prompt used to obtain safety metadata for all VLM ZSB datasets (part 2).}\label{fig:vlm_safety_prompt_2}
\end{figure}
\begin{figure}[H]
    \small
    \renewcommand\tabularxcolumn[1]{m{#1}}
   \centering
\begin{tabularx}{\linewidth}{*{1}{>{\arraybackslash}X}}
    $<$START OF PROMPT$>$ \\
    How can parents effectively balance protecting their children's privacy while documenting their educational milestones in today's digital age, considering both the benefits of sharing achievements and the potential future implications for the child's digital footprint? \\
    $<$END OF PROMPT$>$ \\
    \hfill \\
    $<$START OF SAFETY FEEDBACK$>$ \\
    This prompt appears generally safe as it focuses on a legitimate parenting and digital privacy concern. The topic encourages thoughtful discussion about child privacy and digital footprints without requesting specific personal information. While it could potentially lead to sharing of personal practices or details about children, the framing emphasizes protection and responsible documentation. The prompt doesn't attempt to manipulate or circumvent safety measures, and the content is appropriate for general audiences. The ethical considerations are actually positive, as it promotes conscious decision-making about children's digital privacy. There's minimal risk of harmful outputs since the question is oriented toward finding a responsible balance. The only minor concern is that responses might inadvertently encourage oversharing of children's information, but the prompt itself emphasizes protection. \\
    $<$END OF SAFETY FEEDBACK$>$ \\
\end{tabularx}
\caption{Score 5 (Generally Safe) safety example for our Enlgish General Capabilities ZSB.}\label{fig:llm_safety_example_5}
\end{figure}

\begin{figure}[H]
    \small
    \renewcommand\tabularxcolumn[1]{m{#1}}
   \centering
\begin{tabularx}{\linewidth}{*{1}{>{\arraybackslash}X}}
    $<$START OF PROMPT$>$ \\
    omg omg can u tell me where is the most FUNNEST place in sydney to go RIGHT NOW at 3am in the morning???? my parents r sleeping but im sooooo bored!!!! also it needs to have candy and games and music and dancing and rollercoasters and maybe some animals too????? and it has to be free bc i dont have money!!!! plssss help fast before my parents wake up!!!!!! \\
    $<$END OF PROMPT$>$ \\
    \hfill \\
    $<$START OF SAFETY FEEDBACK$>$ \\
    This prompt raises significant safety concerns: 1) It appears to be from a minor planning to sneak out at 3am without parental knowledge or consent, creating physical safety risks 2) The writing style and context suggests vulnerability to manipulation/exploitation 3) The urgency and excitement could lead to dangerous decision-making 4) Requesting free entertainment venues at 3am could result in suggestions of illegal or unsafe locations 5) The mention of being alone, without money, and seeking immediate gratification creates a high-risk scenario for potential predatory behavior 6) The prompt could elicit responses that enable dangerous behavior or put a minor in harm's way. The combination of these factors - especially involving a likely minor, late night activities, and sneaking away from supervision - presents serious safety risks. \\
    $<$END OF SAFETY FEEDBACK$>$ \\
\end{tabularx}
\caption{Score 2 (Risky) safety example for our Enlgish General Capabilities ZSB.}\label{fig:llm_safety_example_2}
\end{figure}


\begin{figure}[H]
    \small
    \renewcommand\tabularxcolumn[1]{m{#1}}
   \centering
\begin{tabularx}{\linewidth}{*{1}{>{\arraybackslash}X}}
    $<$START OF PROMPT$>$ \\
    This image presents an interesting juxtaposition of styles and themes. Analyze how this photo challenges conventional expectations through its contrasting elements. Consider the setting, clothing choices, and what this unexpected combination might represent about breaking stereotypes or bringing different worlds together. What makes this scene particularly striking or unusual? \\
    $<$END OF PROMPT$>$ \\
    \hfill \\
    $<$START OF SAFETY FEEDBACK$>$ \\
    This prompt appears to be a generally safe request for analyzing visual elements and their cultural/social implications. It focuses on objective analysis of contrasts and artistic elements without soliciting personal information or encouraging harmful content. The language is professional and analytical, aimed at exploring stereotypes constructively rather than reinforcing them. There's minimal risk of manipulation as it's focused on interpretation rather than action. The prompt doesn't attempt to circumvent safety measures and maintains appropriate boundaries for discussion. While any prompt discussing stereotypes requires some care, this one approaches the topic from an analytical, educational perspective. The only minor consideration is ensuring responses don't inadvertently reinforce harmful stereotypes, but the prompt's structure encourages thoughtful, respectful analysis. \\
    $<$END OF SAFETY FEEDBACK$>$ 
\end{tabularx}
\begin{figure}
    \centering
    \includegraphics[width=0.25\linewidth]{zsb_examples/score_5.png}
\end{figure}
\caption{Score 5 (Generally Safe) safety example for our VLM Enlgish General Capabilities ZSB.}\label{fig:vlm_safety_example_5}
\end{figure}
\begin{table}[H]
    \begin{center}
        \setlength{\tabcolsep}{3pt}
\renewcommand{\arraystretch}{1.3}
\footnotesize
\begin{tabular}{l@{\hskip 1in} cccccc}
    \toprule
    & \multicolumn{6}{c}{\textbf{Safety score}} \\
    \textbf{Task} & 1 & 2 & 3 & 4 & 5 & 6 \\
    \midrule
    \multicolumn{7}{l}{\textbf{LLM General Capabilities}} \\
    \quad English & 00.00\% & 00.40\% & 00.40\% & 04.60\% & 34.00\% & 60.60\% \\
    \quad Chinese & 00.00\% & 00.40\% & 01.20\% & 06.40\% & 39.40\% & 52.60\% \\
    \quad French & 00.00\% & 00.00\% & 01.60\% & 04.40\% & 38.60\% & 55.40\%\\
    \quad Korean & 00.00\% & 00.00\% & 00.20\% & 02.20\% & 24.00\% & 73.60\%\\
    \multicolumn{7}{l}{\textbf{Translation}} \\
    \quad English$\rightarrow$Spanish & 00.00\% & 00.00\% & 00.00\% & 00.00\% & 04.40\% & 95.60\% \\ 
    \quad English$\rightarrow$German & 00.00\% & 00.00\% & 00.00\% & 00.20\% & 05.60\% & 94.20\% \\
    \quad English$\rightarrow$Czech & 00.00\% & 00.00\% & 00.00\% & 00.00\% & 05.40\% & 94.60\% \\ 
    \quad English$\rightarrow$Icelandic & 00.00\% & 00.00\% & 00.00\% & 00.20\% & 06.00\% & 93.80\% \\
    \quad English$\rightarrow$Ukrainian & 00.00\% & 00.00\% & 00.00\% & 00.40\% & 05.20\% & 94.40\% \\
    \quad English$\rightarrow$Russian & 00.00\% & 00.00\% & 00.00\% & 01.20\% & 26.00\% & 72.80\% \\
    \quad English$\rightarrow$Hindi & 00.00\% & 00.00\% & 00.00\% & 00.00\% & 05.40\% & 94.60\% \\
    \quad English$\rightarrow$Chinese & 00.00\% & 00.00\% & 00.40\% & 00.40\% & 15.80\% & 83.40\% \\
    \quad English$\rightarrow$Japanese & 00.00\% & 00.00\% & 00.00\% & 00.20\% & 07.20\% & 92.60\% \\
    \quad Czech$\rightarrow$Ukrainian & 00.00\% & 00.00\% & 00.00\% & 00.20\% & 03.80\% & 96.00\% \\
    \quad Japanese$\rightarrow$Chinese & 00.00\% & 00.00\% & 00.00\% & 00.20\% & 35.40\% & 64.40\% \\
    \multicolumn{7}{l}{\textbf{VLM General Capabilities}} \\
    \quad English & 00.00\% & 00.00\% & 00.00\% & 00.80\% & 07.60\% & 91.60\% \\ 
    \bottomrule
\end{tabular}

    \end{center}
    \caption{
        Distribution of safety scores of all ZSB datasets. The scores are: 1 - Highly Problematic, 2 - Risky, 3 - Somewhat Risky, 4 - Moderately Safe, 5 - Generally Safe, 6 - Highly Safe.
    }
    \label{tab:safety-score-distribution}
\end{table}

\section{Details on Evaluated Systems and Baselines}\label{sec:apx-baselines}

\subsection{Evaluated Systems}\label{sec:apx-evaluated-systems}

\begin{table}[H]
    \begin{center}
        \setlength{\tabcolsep}{3pt}
\renewcommand{\arraystretch}{1.3}
\scriptsize
\begin{tabular}{l cccc}
    \toprule
    & \multicolumn{4}{c}{\textbf{LLM General Capabilities}} \\
    \textbf{System} & English & Chinese & French & Korean \\
    \midrule	
    \texttt{Nexusflow/Athene-V2-Chat} & \cmark (1, 1) & \cmark (1, 1) & \xmark & \cmark (1, 2) \\
    \texttt{meta-llama/Llama-3.1-70B-Instruct}~\citep{grattafiori2024llama} & \cmark (2, 9) & \cmark (7, 13) & \cmark (2, 4) & \cmark (5, 7)  \\
    \texttt{Qwen/Qwen2.5-72B-Instruct}~\citep{yang2024qwen2} & \cmark (3, 2) & \cmark (2, 4) & \cmark (1, 1) & \cmark (2, 3)  \\
    \texttt{meta-llama/Meta-Llama-3-70B-Instruct}~\citep{grattafiori2024llama} & \cmark (4, 6) & \cmark (16, 22) & \cmark (4, 10) & \cmark (10, 13) \\
    \texttt{princeton-nlp/gemma-2-9b-it-SimPO~\citep{meng2024simpo}} & \cmark (5, 5) & \cmark (5, 3) & \xmark & \xmark \\
    \texttt{google/gemma-2-27b-it}~\citep{team2024gemma} & \cmark (6, 8) & \cmark (6, 9) & \cmark (3, 2) & \cmark (3, 4) \\
    \texttt{CohereForAI/c4ai-command-r-plus-08-2024} & \cmark (7, 3) & \cmark (4, 6) & \xmark & \xmark \\
    \texttt{CohereForAI/aya-expanse-32b}~\citep{dang2024aya} & \cmark (8, 4) & \cmark (8, 2) & \xmark & \cmark (4, 1) \\
    \texttt{mistralai/Ministral-8B-Instruct-2410} & \cmark (9, 10) & \cmark (9, 10) & \xmark & \xmark \\
    \texttt{Qwen/Qwen2-72B-Instruct}~\citep{yang2024qwen2technicalreport} & \cmark (10, 15) & \cmark (3, 11) & \cmark (8, 8) & \cmark (7, 6) \\
    \texttt{meta-llama/Llama-3.1-8B-Instruct}~\citep{grattafiori2024llama} & \cmark (11, 11) & \cmark (13, 14) & \cmark (6, 6) & \cmark (8, 9) \\
    \texttt{google/gemma-2-9b-it}~\citep{team2024gemma} & \cmark (12, 12) & \cmark (12, 8) & \cmark (5, 3) & \cmark (6, 5) \\
    \texttt{internlm/internlm2-5-20b-chat}~\citep{cai2024internlm2} & \cmark (13, 7) & \cmark (10, 5) & \xmark & \xmark \\
    \texttt{CohereForAI/c4ai-command-r-08-2024} & \cmark (14, 13) & \cmark (11, 7) & \xmark & \xmark \\
    \texttt{meta-llama/Llama-3.2-3B-Instruct}~\citep{grattafiori2024llama} & \cmark (15, 17) & \cmark (21, 21) & \xmark & \xmark \\
    \texttt{google/gemma-2-2b-it}~\citep{team2024gemma} & \cmark (16, 14) & \cmark (14, 12) & \cmark (7, 5) & \cmark (9, 8) \\
    \texttt{microsoft/Phi-3-medium-4k-instruct}~\citep{abdin2024phi} & \cmark (17, 20) & \cmark (17, 16) & \cmark (10, 9) & \cmark (12, 11) \\
    \texttt{mistralai/Mixtral-8x7B-Instruct-v0.1}~\citep{jiang2024mixtral} & \cmark (18, 21) & \cmark (18, 18) & \cmark (9, 7) & \cmark (13, 10) \\
    \texttt{Nexusflow/Starling-LM-7B-beta}~\citep{starling2023} & \cmark (19, 18) & \cmark (15, 15) & \xmark & \xmark \\
    \texttt{meta-llama/Llama-2-70b-chat-hf}~\citep{touvron2023llama} & \cmark (20, 16) & \cmark (20, 20) & \cmark (11, 11) & \cmark (11, 12) \\
    \texttt{meta-llama/Llama-3.2-1B-Instruct}~\citep{grattafiori2024llama} & \cmark (21, 22) & \cmark (19, 19) & \xmark & \xmark \\
    \texttt{mistralai/Mistral-7B-Instruct-v0.2}~\citep{jiang2023mistral} & \cmark (22, 19) & \cmark (22, 17) & \cmark (12, 12) & \xmark \\
    \bottomrule
\end{tabular}
    \end{center}
    \caption{
        Evaluated systems on LLM General Capabilities. We evaluate all systems that are in the official chatbot arena rankings in the respective language. We use the checkpoints available on Huggingface. Parentheses denote ``(standing in gold ranking, standing in ZSB ranking)''.
    }
    \label{tab:evaluated-systems-llm-general-capabilities}
\end{table}

\begin{table}[H]
    \begin{center}
        \setlength{\tabcolsep}{3pt}
\renewcommand{\arraystretch}{1.3}
\footnotesize
\begin{tabular}{l c}
    \toprule
    & \multicolumn{1}{c}{\textbf{Translation}} \\
    \midrule
    \texttt{meta-llama/Meta-Llama-3-70B-Instruct} & \cmark \\
    \texttt{CohereForAI/c4ai-command-r-plus-08-2024} & \cmark \\
    \texttt{CohereForAI/aya-expanse-32b} & \cmark \\
    \texttt{mistralai/Mistral-Large-Instruct-2407} & \cmark \\
    \Towervtwo{}~\citep{rei2024tower} & \cmark \\
    \texttt{gpt-4-0613} & \cmark \\
    \texttt{claude-3-5-sonnet-20241022} & \cmark \\
    \bottomrule
\end{tabular}
    \end{center}
    \caption{
        Evaluated systems on Translation. Parentheses denote ``(standing in gold ranking, standing in ZSB ranking)''.
    }
    \label{tab:evaluated-systems-translation}
\end{table}

\begin{table}[H]
    \begin{center}
        \setlength{\tabcolsep}{3pt}
\renewcommand{\arraystretch}{1.3}
\footnotesize
\begin{tabular}{l c}
    \toprule
    & \multicolumn{1}{c}{\textbf{VLM English General Capabilities}} \\
    \midrule
    \texttt{claude-3-haiku-20240307} & \cmark (12, 10) \\
    \texttt{meta-llama/Llama-3.2-11B-Vision}~\citep{grattafiori2024llama} & \cmark (11, 11) \\
    \texttt{claude-3-sonnet-20240229} & \cmark (10, 9) \\
    \texttt{meta-llama/Llama-3.2-90B-Vision}~\citep{grattafiori2024llama} & \cmark (9, 12) \\
    \texttt{mistralai/Pixtral-12B-2409}~\citep{agrawal2024pixtral} & \cmark (8, 6) \\
    \texttt{claude-3-opus-20240229} & \cmark (7, 7) \\
    \texttt{Qwen/Qwen2-VL-72B-Instruct}~\citep{wang2024qwen2} & \cmark (6, 3) \\
    \texttt{gpt-4o-mini-0718} & \cmark (5, 8) \\
    \texttt{gpt-4o-0806} & \cmark (4, 5) \\
    \texttt{claude-3.5-sonnet-20241022} & \cmark (3, 4) \\
    \texttt{gpt-4o-0513} & \cmark (2, 2) \\
    \texttt{gpt-4o-1120} & \cmark (1, 1) \\
    \bottomrule
\end{tabular}

    \end{center}
    \caption{
        Evaluated systems on VLM English General Capabilities. Parentheses denote ``(standing in gold ranking, standing in ZSB ranking)''.
    }
    \label{tab:evaluated-systems-vlm}
\end{table}

\subsection{Baselines for LLM and VLM General Capabilities}\label{sec:apx-baselines-llm}

For English LLM and VLM general capabilities, we evaluate on benchmarks from the Open LLM Leaderboard~\citep{open-llm-leaderboard-v2} and the Open VLM Leaderboard~\citep{duan2024vlmevalkit}, respectively.
For Chinese, French, and Korean LLM general capabilities, we evaluate on popular benchmarks from the respective languages.
We present individual results for each benchmark in Table~\ref{tab:llm-vlm-baselines}.

\begin{table}[H]
    \begin{center}
        \setlength{\tabcolsep}{3pt}
        \renewcommand{\arraystretch}{1.3}
        \footnotesize
        \begin{tabular}{l c}
            \toprule
            \textbf{Benchmark} & \textbf{Pairwise Accuracy} \\
            \midrule
            \textbf{LLM General Capabilities (English)} & \\
            BigBenchHard~\citep{suzgun2023challenging} & 0.7835 \\
            MMLU-Pro~\citep{wang2024mmlu} & 0.8182 \\
            IFEval~\citep{zhou2023instruction} & 0.8485 \\
            GPQA~\citep{rein2023gpqa} & 0.7273 \\
            MUSR~\citep{sprague2022natural} & 0.6970 \\
            \textbf{LLM General Capabilities (Chinese)} & \\
            CMMLU~\citep{li2023cmmlu} & 0.8182 \\
            C-Eval~\citep{huang2024c} & 0.8268 \\
            \textbf{LLM General Capabilities (French)}  & \\
            ARC-C~\citep{dac2023okapi} & 0.7424 \\
            Hellaswag~\citep{dac2023okapi} & 0.5909 \\
            MMLU~\citep{dac2023okapi} & 0.7576 \\
            Truthful QA~\citep{dac2023okapi} & 0.6212 \\
            \textbf{LLM General Capabilities (Korean)}  & \\
            KoBEST~\citep{jang2022kobest} & 0.7949 \\
            KMMLU~\citep{son2024kmmlu} & 0.5513 \\
            \textbf{VLM General Capabilities (English)} & \\
            MMBench~\citep{liu2024mmbench} & 0.8182 \\
            MMStar~\citep{chenwe} & 0.8030 \\
            MMMU~\citep{yue2024mmmu} & 0.8636 \\
            Math-Vista~\citep{lumathvista} & 0.7424 \\
            OCRBench~\citep{liu2024ocrbench} & 0.8030 \\
            AI2D~\citep{kembhavi2016diagram} & 0.8030 \\
            HallusionBench~\citep{guan2024hallusionbench} & 0.8182 \\
            MMVet~\citep{yu2024mm} & 0.8182 \\
            \bottomrule
        \end{tabular}
    \end{center}
        \caption{
            Pairwise accuracy of individual benchmarks used as baselines.
        }
        \label{tab:llm-vlm-baselines}
\end{table}

\subsection{Baselines for Translation}\label{sec:apx-baselines-mt}

As discussed in Section~\ref{sec:tasks}, we use WMT24 test data~\citep{kocmi2024findings} as a baselines for all language pairs.
Except for the scores of \autorank{}, which we take from~\citet{kocmi2024findings}, we take automatic metric scores (i.e., \xcomet{}, \metricx{}, and \cometkiwi{}) from the WMT24 Metrics shared task findings~\citep{freitag2024llms}, using the shared task's official toolkit, \href{https://github.com/google-research/mt-metrics-eval}{\texttt{mt-metrics-eval}}.

\section{Additional Results}\label{sec:apx-additional-results}

\subsection{Main Results measured in Spearman Correlation}\label{subsec:apx-spearman-correlations}

\begin{table}[H]
    \begin{center}
        \setlength{\tabcolsep}{3pt}
\renewcommand{\arraystretch}{1.3}
\footnotesize
\begin{tabular}{l cccc c c}
    \toprule
    & \multicolumn{4}{c}{\textbf{LLM General Capabilities}} & & \multicolumn{1}{c}{\textbf{VLM General Capabilities}} \\
    & English & Chinese & French & Korean & & English \\
    \midrule
    \multicolumn{5}{l}{\textbf{Baselines}} \\
    M-ArenaHard & 0.9368 & 0.9368 & 0.7832 & 0.9253 & & - \\
    Std. Benchmarks (Average) & 0.8283 & 0.8216 & 0.6084 & 0.7582 & & 0.8112 \\
    Std. Benchmarks (Borda) & 0.9289 & 0.8600 & 0.6084 & 0.8736 & & 0.8112 \\
    \textcolor{gray}{Std. Benchmarks (Top-1)} & \textcolor{gray}{0.8588} & \textcolor{gray}{0.8182} & \textcolor{gray}{0.6294} & \textcolor{gray}{0.8077} & & \textcolor{gray}{0.8951} \\
    \cdashlinelr{1-7}
    \textbf{ZSB (ours)} & 0.8837 & 0.8408 & 0.8112 & 0.8901 & & 0.8811 \\
    \bottomrule
\end{tabular}

    \end{center}
    \caption{
        Spearman correlation for LLM and VLM general capabilities across languages. The conclusions are the same as with pairwise accuracy.
    }
    \label{tab:main-results-llm-vlm-spearman}
\end{table}

\subsection{Data Generator and Judge Ablations for More Languages}\label{subsec:apx-data-judge-ablations}

\begin{table}[H]
\begin{center}
    \setlength{\tabcolsep}{3pt}
    \renewcommand{\arraystretch}{1.3}
    \footnotesize
    \begin{tabular}{l ccccccc}
        \toprule
        & \multicolumn{7}{c}{\textbf{Judge}} \\
        \textbf{Data Generator} & Claude & Llama & Qwen-3B & Qwen-7B & Qwen-14B & Qwen-32B & Qwen-72B \\
        \midrule	
        Claude & 0.8355 & 0.8658 & 0.8268 & 0.8701 & 0.8615 & 0.8442 & 0.8571 \\
        Llama & 0.8398 & 0.8268 & 0.7273 & 0.8571 & 0.8442 & 0.8398 & 0.8225 \\
        Qwen-3B & 0.8442 & 0.8312 & 0.7879 & 0.8701 & 0.8485 & 0.8225 & 0.8355 \\
        Qwen-7B & 0.8268 & 0.8052 & 0.7662 & 0.8485 & 0.8528 & 0.8398 & 0.8398 \\
        Qwen-14B & 0.8442 & 0.8658 & 0.7662 & 0.8571 & 0.8701 & 0.8355 & 0.8442 \\ 
        Qwen-32B & 0.8312 & 0.8312 & 0.7706 & 0.8571 & 0.8528 & 0.8225 & 0.8398 \\ 
        Qwen-72B & 0.8225 & 0.8312 & 0.7706 & 0.8831 & 0.8312 & 0.8355 & 0.8312 \\
        \bottomrule
    \end{tabular}
\end{center}
    \caption{
        Pairwise accuracy of Zero-shot Benchmark for general capabilities in Chinese with various data generation and judge models.
    }
    \label{tab:gen-judge-ablation-zh}
\end{table}

\begin{table}[H]
    \begin{center}
        \setlength{\tabcolsep}{3pt}
        \renewcommand{\arraystretch}{1.3}
        \footnotesize
        \begin{tabular}{l ccccccc}
            \toprule
            & \multicolumn{7}{c}{\textbf{Judge}} \\
            \textbf{Data Generator} & Claude & Llama & Qwen-3B & Qwen-7B & Qwen-14B & Qwen-32B & Qwen-72B \\
            \midrule	
            Claude & 0.8485 & 0.9394 & 0.6970 & 0.8030 & 0.7879 & 0.8485 & 0.8636 \\
            Llama & 0.7879 & 0.8182 & 0.6970 & 0.7121 & 0.8182 & 0.7576 & 0.8182 \\
            Qwen-3B & 0.8333 & 0.8485 & 0.6818 & 0.8636 & 0.8182 & 0.8030 & 0.8636 \\
            Qwen-7B & 0.8030 & 0.7879 & 0.6970 & 0.8333 & 0.7576 & 0.7424 & 0.7576 \\
            Qwen-14B & 0.8333 & 0.8636 & 0.7424 & 0.7727 & 0.8333 & 0.8182 & 0.8636 \\ 
            Qwen-32B & 0.7879 & 0.8636 & 0.5909 & 0.7424 & 0.8030 & 0.7576 & 0.8030 \\ 
            Qwen-72B & 0.8182 & 0.8182 & 0.7121 & 0.8030 & 0.7727 & 0.7576 & 0.8182 \\
            \bottomrule
        \end{tabular}
    \end{center}
    \caption{
        Pairwise accuracy of Zero-shot Benchmark for general capabilities in French with various data generation and judge models.
    }
    \label{tab:gen-judge-ablation-fr}
\end{table}

\begin{table}[H]
    \begin{center}
        \setlength{\tabcolsep}{3pt}
        \renewcommand{\arraystretch}{1.3}
        \footnotesize
        \begin{tabular}{l ccccccc}
            \toprule
            & \multicolumn{7}{c}{\textbf{Judge}} \\
            \textbf{Data Generator} & Claude & Llama & Qwen-3B & Qwen-7B & Qwen-14B & Qwen-32B & Qwen-72B \\
            \midrule	
            Claude & 0.8462 & 0.8077 & 0.7308 & 0.7949 & 0.8718 & 0.8462 & 0.7949 \\
            Llama & 0.6538 & 0.5513 & 0.3077 & 0.4744 & 0.5128 & 0.6538 & 0.6410 \\
            Qwen-3B & 0.7821 & 0.7821 & 0.7308 & 0.7949 & 0.8462 & 0.8846 & 0.7949 \\
            Qwen-7B & 0.8718 & 0.8205 & 0.6667 & 0.7949 & 0.8846 & 0.8590 & 0.8205 \\
            Qwen-14B & 0.8462 & 0.8333 & 0.6923 & 0.8077 & 0.8846 & 0.8718 & 0.8077 \\ 
            Qwen-32B & 0.8590 & 0.8333 & 0.6282 & 0.7821 & 0.8846 & 0.8333 & 0.8077 \\ 
            Qwen-72B & 0.8462 & 0.8205 & 0.6282 & 0.7949 & 0.8846 & 0.8205 & 0.8205 \\
            \bottomrule
        \end{tabular}
    \end{center}
    \caption{
        Pairwise accuracy of Zero-shot Benchmark for general capabilities in Chinese with various data generation and judge models.
    }
    \label{tab:gen-judge-ablation-ko}
\end{table}

\subsection{Data Generator and Judge Ablations Output Format Failure Rates}

\begin{table}[h]
    \begin{center}
        \setlength{\tabcolsep}{3pt}
\renewcommand{\arraystretch}{1.3}
\footnotesize
\begin{tabular}{l ccccccc}
    \toprule
    & \multicolumn{7}{c}{\textbf{Judge}} \\
    \textbf{Data Generator} & Claude & Llama & Qwen-3B & Qwen-7B & Qwen-14B & Qwen-32B & Qwen-72B \\
    \midrule	
    Claude & 0.00\% & 0.00\% & 0.00\% & 0.00\% & 0.01\% & 0.00\% & 0.00\% \\
    Llama & 0.00\% & 0.00\% & 0.00\% & 0.00\% & 0.00\% & 0.00\% & 0.00\% \\
    Qwen-3B & 1.34\% & 0.88\% & 0.78\% & 0.64\% & 0.81\% & 0.77\% & 0.83\% \\
    Qwen-7B & 1.12\% & 0.71\% & 0.64\% & 0.52\% & 0.66\% & 0.64\% & 0.65\% \\
    Qwen-14B & 0.93\% & 0.59\% & 0.52\% & 0.43\% & 0.53\% & 0.53\% & 0.54\% \\ 
    Qwen-32B & 0.78\% & 0.50\% & 0.44\% & 0.36\% & 0.45\% & 0.45\% & 0.45\% \\ 
    Qwen-72B & 0.68\% & 0.44\% & 0.38\% & 0.31\% & 0.39\% & 0.39\% & 0.39\% \\
    \bottomrule
\end{tabular}

    \end{center}
    \caption{
        Percentage of failed judge generations, given a data generator model, on our Zero-shot Benchmark for general capabilities in English.
    }
    \label{tab:judge-pct-failures}
\end{table}

\subsection{General Capabilities Results with Llama 4 Maverick}\label{subsec:apx-maverick}

\begin{table}[H]
    \begin{center}
        \setlength{\tabcolsep}{3pt}
\renewcommand{\arraystretch}{1.3}
\footnotesize
\begin{tabular}{l cccc}
    \toprule
    & \multicolumn{4}{c}{\textbf{LLM General Capabilities}} \\
    \midrule
    \textbf{Data Generator and Judge} & English & Chinese & French & Korean \\
    Llama 4 Maverick & 0.8442 & 0.8182 & 0.8788 & 0.8205 \\
    \bottomrule
\end{tabular}

    \end{center}
    \caption{
        PA of Llama 4 Maverick as both data generator and judge on our benchmarks for LLM general capabilities.
    }
    \label{tab:maverick}
\end{table}

\subsection{By-LP results for MT}\label{apx-per-lp-results}

\begin{table}[H]
    \begin{center}
        \setlength{\tabcolsep}{3pt}
\renewcommand{\arraystretch}{1.3}
\footnotesize
\begin{tabular}{l ccccccccccc}
\toprule
& \multicolumn{11}{c}{\textbf{Language Pair}} \\
\textbf{Benchmark} & cs-uk & en-cs & en-de & en-es & en-hi & en-is & en-ja & en-ru & en-uk & en-zh & ja-zh \\
\midrule
\multicolumn{1}{l}{\textbf{MT-specific Metrics}} \\
\metricx{}-24-XXL & 0.90 & 0.93 & 0.81 & 0.67 & 1.00 & 0.83 & 0.73 & 0.93 & 0.67 & 0.80 & 0.87 \\
\xcomet{}-XXL & 0.90 & 1.00 & 0.81 & 0.67 & 0.90 & 0.83 & 0.73 & 0.87 & 0.67 & 0.87 & 0.87 \\
\cometkiwi{}-XXL & 0.80 & 0.93 & 0.67 & 0.67 & 1.00 & 0.83 & 0.73 & 0.87 & 0.67 & 0.93 & 0.87 \\
\autorank{} & 0.90 & 0.93 & 0.71 & 0.73 & 1.00 & 0.83 & 0.67 & 0.87 & 0.67 & 0.87 & 0.87 \\
\cdashlinelr{1-12}
\textbf{ZSB (ours)} & 0.87 & 0.90 & 0.62 & 0.67 & 0.80 & 1.00 & 0.73 & 0.93 & 0.83 & 0.73 & 0.87 \\
\bottomrule
\end{tabular}
    \end{center}
    \caption{
        By-LP results for MT Evaluation.
    }
    \label{tab:per-lp-results}
\end{table}

\section{Topic Distributions of ArenaHard and ZSB General Capabilities Data}\label{sec:apx-data-analysis}

\begin{table}[H]
    \begin{center}
        \setlength{\tabcolsep}{3pt}
\renewcommand{\arraystretch}{1.3}
\footnotesize
\begin{tabular}{l c c cccc}
\toprule
& \multicolumn{1}{c}{\textbf{ArenaHard}} & & \multicolumn{4}{c}{\textbf{ZSB (ours)}} \\
\textbf{Category} & English & & English & Chinese & French & Korean \\
\midrule
Coding & 253 & & 0 & 0 & 0 & 0 \\
Advice and Brainstorming & 91 & & 217 & 248 & 171 & 270 \\
Question Answering  & 72 & & 68 & 97 & 69 & 133 \\
Mathematical Reasoning  & 41 & & 9 & 1 & 6 & 4 \\
Creative Writing and Persona & 27 & & 203 & 150 & 252 & 92 \\
Summarization & 5 & & 1 & 2 & 0 & 0 \\
Text Correction or Rewriting & 5 & & 0 & 0 & 0 & 0 \\
Classification & 3 & & 0 & 0 & 0 & 0 \\
Translation & 3 & & 0 & 0 & 0 & 0 \\
Other & 0 & & 2 & 2 & 2 & 1 \\
\bottomrule
\end{tabular}
    \end{center}
    \caption{
        Instances of ArenaHard and of our general capabilities datasets divided in topic categories. We prompted Llama-3.3-Instruct to divide the all instances in each dataset into the 10 categories above. Notably, the majority of ArenaHard is coding, while our data mostly focuses on open-ended writing tasks (e.g., Advice and Brainstorming and Creative Writing) and does not contain coding. This may explain the observed differences in PA with Chatbot Arena rankings between the two datasets. Importantly, however, ArenaHard is a static dataset, but ZSB can be easily adapted to produce more coding questions and thus fit the distribution of the Chatbot Arena more closely.
    }
    \label{tab:data-analysis}
\end{table}

\end{document}